\documentclass[a4paper,fleqn]{cas-sc}

\usepackage[numbers]{natbib}
\usepackage{float}
\usepackage{placeins}
\usepackage{subcaption}
\usepackage{graphicx}
\usepackage{caption}
\usepackage{makecell}
\usepackage{etoolbox}
\usepackage{xspace}
\usepackage{adjustbox}
\usepackage{capt-of}
\usepackage{tikz}
\usepackage{float}
\usepackage{adjustbox}
\usetikzlibrary{
    arrows.meta,
    positioning,
    shapes.geometric,
    fit,
    calc,
    backgrounds
}

\def\tsc#1{\csdef{#1}{\textsc{\lowercase{#1}}\xspace}}
\tsc{Reload}
\usetikzlibrary{arrows.meta, positioning, shapes.geometric, fit}
\begin{document}
\let\WriteBookmarks\relax

\shorttitle{Reload-Mamba}
\shortauthors{S.W. Chan}

\title [mode = title]{Reload-Mamba: Hierarchical Anti-Dilution State-Space Modeling for Multi-Class Semantic Segmentation}

\author[1]{Sheng-Wei Chan}[orcid=0009-0002-3983-5163]
\author[1]{Hsin-Jui Pan}
\author[1]{Jen-Shiun Chiang\corref{cor1}}
\ead{chiang@mail.tku.edu.tw}

\affiliation[1]{
  organization={Department of Electrical and Computer Engineering, Tamkang University},
  addressline={No.151, Yingzhuan Rd.},
  city={Tamsui Dist., New Taipei City},
  postcode={251301},
  country={Taiwan}
}

\cortext[cor1]{Corresponding author}

\begin{abstract}
Mamba-based state space models offer linear-time long-range modeling for high-resolution dense prediction, but sequential state-space propagation can attenuate boundary-sensitive and detail-sensitive responses that are critical in multi-class semantic segmentation. We propose Reload-Mamba, a semantic segmentation framework that addresses this propagation-induced response dilution through three segmentation-specific designs: (i) a \textbf{boundary-supervised local detail prior} that is explicitly trained with ground-truth boundary masks to identify regions requiring response restoration; (ii) a \textbf{class-uncertainty-aware Reload Gate} that incorporates per-pixel class entropy from a pre-reload auxiliary head as an additional gating signal, a formulation that is informative only under multi-class dense prediction; and (iii) a \textbf{hierarchical multi-level Reload} mechanism that applies anti-dilution refinement at three decoder levels and fuses the restored representations top-down. Built upon a ConvNeXt-Tiny encoder with a multi-scale decoder and four-directional Mamba scanning with pixel-wise directional attention, Reload-Mamba achieves \textbf{47.9\%} single-scale (\textbf{48.9\%} multi-scale) mIoU on ADE20K and \textbf{83.2\%} single-scale mIoU on Cityscapes. With ResNet-101 + COCO pre-training under the standard DeepLab-style protocol, Reload-Mamba reaches \textbf{87.8\%} mIoU on PASCAL VOC 2012 val. Controlled ablations show that each of the three segmentation-specific designs contributes beyond a direct port of the prior anti-dilution architecture proposed for binarization, cumulatively improving over the direct-port baseline by $+2.2$ mIoU on ADE20K.
\end{abstract}


\maketitle

\section{Introduction}

Semantic segmentation is a fundamental dense prediction task that assigns a semantic label to each pixel \cite{long2015fcn,ronneberger2015unet,chen2018deeplabv3plus}. Unlike image-level recognition, semantic segmentation requires both global semantic understanding and local spatial precision. In particular, object boundaries, thin structures, and small semantic regions often rely on subtle local responses. Therefore, an effective segmentation model should balance long-range dependency modeling and fine-grained detail preservation. Recently, Mamba-based state space models have attracted increasing attention in vision tasks due to their efficient linear-time sequence modeling capability \cite{gu2023mamba,zhu2024vim,liu2024vmamba}. Compared with Transformer-based self-attention, Mamba provides a computationally efficient way to model long-range dependencies, making it attractive for high-resolution dense prediction \cite{dosovitskiy2021vit,liu2021swin,xie2021segformer}. However, direct state-space propagation may weaken local detail responses because features are sequentially mixed along scan paths. In semantic segmentation, such response attenuation may lead to blurred object boundaries, missed thin structures, or reduced sensitivity to small semantic regions. Sequential state-space propagation in Mamba-based dense prediction
can attenuate boundary-sensitive and detail-sensitive responses, an
effect we refer to as propagation-induced response dilution. In our
prior work~\cite{chan2026deepmine}, we identified this phenomenon in
the context of document image binarization and proposed a
single-level anti-dilution refinement that restores weakened
foreground responses through a learned gate. While that formulation
is effective for binary stroke restoration, multi-class semantic
segmentation imposes three additional requirements that a direct
port of the single-level binarization design cannot satisfy. First, the regions that benefit from response
restoration---class boundaries, thin structures, and small
categories---are not uniquely identifiable from class probabilities
alone, so an implicitly estimated detail prior provides only weak
conditioning. Second, ambiguous pixels concentrate at semantic
transitions and are best identified via per-pixel class entropy, a
signal that is uninformative under binary prediction but
discriminative when the number of classes exceeds two. Third,
dilution affects different semantic scales differently, and
refinement at a single decoder resolution cannot simultaneously serve
large coherent regions, mid-scale objects, and thin boundary
structures.

In this paper, we propose Reload-Mamba, a semantic segmentation framework that addresses propagation-induced response dilution through three segmentation-specific designs. First, we explicitly supervise the local detail prior using ground-truth boundary masks at every decoder level, replacing the implicit prior estimation with a stable conditioning signal that reliably localizes detail-sensitive regions. Second, we attach a lightweight pre-reload auxiliary head to the directional fusion output and feed its per-pixel class entropy into the Reload Gate as a third input alongside the original feature and the propagation-induced feature change; this entropy term is fundamentally specific to multi-class prediction and provides a complementary signal to the prior. Third, we apply the Reload Gate at three decoder levels and fuse the restored representations top-down, so that anti-dilution refinement operates at the resolution most appropriate to each semantic scale. With a ConvNeXt-Tiny backbone, Reload-Mamba obtains 47.9\% single-scale mIoU on ADE20K and 83.2\% single-scale mIoU on Cityscapes; with ResNet-101 + COCO pre-training, it reaches 87.8\% on PASCAL VOC val. Controlled ablations confirm that each of the three designs contributes beyond a direct port of the prior anti-dilution architecture.

The main contributions of this paper are summarized as follows:
\begin{itemize}
\item We identify three concrete limitations that prevent a direct port of binarization-style anti-dilution from being effective in multi-class semantic segmentation, and propose Reload-Mamba, a segmentation framework whose hierarchical anti-dilution refinement is applied at three decoder levels rather than a single feature resolution.

\item We design a class-uncertainty-aware Reload Gate that jointly considers the original decoder feature, the propagation-induced feature change, and the per-pixel class uncertainty derived from a pre-reload auxiliary head. This formulation is informative only when the number of classes exceeds two, and is therefore fundamentally specific to multi-class dense prediction.

\item We introduce an explicit boundary-supervised local detail prior that uses ground-truth boundary masks to guide prior estimation across multiple decoder levels, providing a more reliable conditioning signal than the implicit prior used in the original binarization formulation.

\item We conduct extensive experiments on ADE20K, Cityscapes, and PASCAL VOC. Reload-Mamba reaches 47.9\% single-scale (48.9\% multi-scale) mIoU on ADE20K, 83.2\% on Cityscapes, and 87.8\% on PASCAL VOC val. Controlled ablations show that the three segmentation-specific designs cumulatively contribute $+2.2$ mIoU on ADE20K beyond a direct port of the prior anti-dilution architecture, and the resulting model matches or surpasses strong attention-based and dilation-based segmentation methods under comparable settings.
\end{itemize}

\section{Related Work}

\subsection{Semantic Segmentation}
Semantic segmentation is a fundamental dense prediction task that assigns a semantic label to each pixel in an image. Early deep learning-based approaches established the fully convolutional formulation by replacing fully connected layers with convolutional prediction heads, enabling end-to-end pixel-wise classification \cite{long2015fcn}. Following this direction, encoder-decoder architectures became widely adopted because they can combine high-level semantic representations with low-level spatial details. U-Net introduced a symmetric encoder-decoder structure with skip connections to recover fine-grained localization information \cite{ronneberger2015unet}. Later segmentation models further improved context aggregation and boundary recovery. PSPNet exploited pyramid pooling to incorporate global scene-level context \cite{zhao2017pspnet}, while DeepLab series adopted atrous convolution and atrous spatial pyramid pooling to enlarge the effective receptive field without severely reducing spatial resolution \cite{chen2017deeplab,chen2018deeplabv3plus}. DeepLabv3+ further combined atrous spatial pyramid pooling with a decoder module, improving object boundary refinement \cite{chen2018deeplabv3plus}. Recent semantic segmentation methods increasingly rely on strong hierarchical backbones and an efficient decoding head. Transformer-based segmentation methods exploit self-attention for global context modeling, leading to strong performance in complex scenes \cite{dosovitskiy2021vit,liu2021swin,xie2021segformer}. Swin Transformer introduced shifted-window attention to reduce the cost of global self-attention while preserving hierarchical representations for dense prediction \cite{liu2021swin}. SegFormer further simplified the segmentation pipeline by using a hierarchical Transformer encoder and an efficient MLP decoder, showing strong accuracy-efficiency trade-offs on ADE20K and Cityscapes \cite{xie2021segformer}. Despite their effectiveness, attention-based models often suffer from quadratic complexity with respect to token length, which motivates the exploration of more efficient long-range modeling mechanisms.

\subsection{Efficient Backbones for Dense Prediction}
Backbone design is critical for semantic segmentation because dense prediction requires both strong semantic abstraction and accurate spatial localization. Classical convolutional backbones such as VGG, ResNet, and ResNeXt have been widely used in segmentation frameworks due to their local inductive bias and efficient feature extraction \cite{simonyan2015vgg,he2016resnet,xie2017resnext}. More recently, ConvNeXt revisited the design of convolutional networks under modern training and architectural principles, demonstrating that pure ConvNets can remain competitive with Transformer-based backbones while preserving simplicity and efficiency \cite{liu2022convnext}. This makes ConvNeXt a suitable backbone for segmentation systems that require stable multi-scale representations and efficient dense feature extraction. In dense prediction, multi-scale feature fusion is commonly used to recover spatial resolution and integrate semantic information across different stages \cite{lin2017fpn,chen2018deeplabv3plus,xie2021segformer}. Skip connections and decoder modules are especially important for restoring details that may be lost in deep hierarchical encoders \cite{ronneberger2015unet,badrinarayanan2017segnet,chen2018deeplabv3plus}. Our work follows this principle by adopting a ConvNeXt-Tiny encoder and a multi-scale decoder to produce a compact yet effective segmentation backbone. Different from existing decoder-centric refinement strategies, our method introduces a Mamba-based directional context module together with hierarchical anti-dilution refinement to address the feature dilution problem caused by sequential state modeling.

\subsection{State Space Models and Vision Mamba}
State space models have recently attracted increasing attention as efficient alternatives to self-attention for long-sequence modeling. Structured state space models such as S4 demonstrated the potential of sequence modeling with long-range dependency capture and subquadratic complexity \cite{gu2022s4}. Mamba further improved this line of research by introducing selective state spaces and a hardware-aware implementation, enabling input-dependent sequence modeling with linear-time complexity \cite{gu2023mamba}. Compared with Transformers, Mamba provides an efficient mechanism for modeling long-range dependencies, which is attractive for high-resolution vision tasks where token lengths can become very large. Several works have extended Mamba and state space models to vision tasks. Vision Mamba (Vim) introduced bidirectional state space modeling for visual representation learning and treated image patches as sequences, showing that SSM-based visual backbones can serve as efficient alternatives to attention-based models \cite{zhu2024vim}. VMamba further proposed a visual state space backbone with 2D selective scanning, adapting Mamba to image data through spatial traversal strategies \cite{liu2024vmamba}. These works show that scan direction and spatial ordering are crucial when applying sequence models to two-dimensional images. Other studies have explored multi-scale or hierarchical visual Mamba designs to improve the trade-off between efficiency and accuracy \cite{shi2024msvmamba,ruan2024vm-unet}. Mamba-based models have also been investigated for segmentation. SegMamba introduced Mamba for 3D medical image segmentation to capture long-range dependencies in volumetric features \cite{xing2024segmamba}. U-Mamba combined convolutional operations with Mamba blocks in an encoder-decoder segmentation framework, aiming to preserve local feature extraction while improving long-range dependency modeling \cite{ma2024umamba}. Our prior work~\cite{chan2026deepmine} examined response dilution under selective scanning in the binarization setting and proposed a single-level anti-dilution refinement to restore weakened stroke responses. While that formulation is effective for binary foreground restoration, the segmentation-specific designs and multi-level refinement required for multi-class dense prediction were not addressed in that study. The present work extends this line of investigation by reformulating anti-dilution refinement for multi-class semantic segmentation, where boundary identification, class ambiguity, and multi-scale dilution all need to be handled explicitly. Other studies demonstrate the potential of Mamba for dense prediction more broadly, especially when global context is required, but directly applying sequential scanning to segmentation may weaken local detail responses because features are repeatedly propagated and mixed along scan paths.

\subsection{Detail Restoration and Boundary-Aware Refinement}

Accurate semantic segmentation requires not only global context but also fine boundary localization. Many methods therefore introduce boundary-aware learning, auxiliary supervision, or detail refinement modules to improve segmentation quality near object contours \cite{chen2018deeplabv3plus,takikawa2019gatedscnn,ding2019boundary,ke2020sfnet}. Gated-SCNN explicitly used shape information and a gating mechanism to refine boundaries \cite{takikawa2019gatedscnn}, while other approaches explored semantic-flow alignment or boundary supervision to recover spatial details from high-level features \cite{ding2019boundary,ke2020sfnet}. Auxiliary heads and deep supervision are also commonly used to stabilize intermediate representations and improve gradient propagation in segmentation networks \cite{lee2015deeply,zhao2017pspnet,xie2021segformer}. Edge priors provide another effective way to enhance local structure. Classical operators such as Sobel filters capture gradient-based boundary cues and can be incorporated as additional detail information for segmentation refinement \cite{sobel1968operator}. In modern deep segmentation models, edge-aware branches or boundary losses are often used to emphasize difficult pixels around class transitions \cite{takikawa2019gatedscnn,ding2019boundary,chen2018deeplabv3plus}. In this work, we introduce a Sobel edge prior and a detail fusion branch to enhance boundary-aware representations. More importantly, we propose a hierarchical anti-dilution refinement mechanism to recover local responses after directional Mamba scanning at multiple decoder levels. This design complements existing boundary refinement strategies by addressing a specific limitation of Mamba-based segmentation: the multi-scale dilution of local details during long-range sequential propagation.

\section{Overall Model Architecture}
The overall architecture of Reload-Mamba is shown in Figure~\ref{fig:reload_mamba_arch}. The proposed network follows an encoder-decoder segmentation paradigm and introduces a hierarchical directional state-space modeling module together with a class-uncertainty-aware local detail restoration mechanism. Given an input image $I \in \mathbb{R}^{3 \times H \times W}$, the network first extracts hierarchical visual representations using ConvNeXt-Tiny as the encoder backbone. These features are then passed to a lightweight multi-scale decoder, which fuses shallow spatial features and deep semantic features through top-down feature aggregation. After that, the proposed hierarchical Reload-Mamba module is applied at three decoder levels ($D_4$, $D_3$, $D_2$) to perform directional state-space context modeling and class-uncertainty-aware anti-dilution refinement at multiple semantic scales. Finally, edge-guided detail fusion and a segmentation head are used to produce the final prediction map.

\noindent
\begin{minipage}{\linewidth}
\centering

\begin{adjustbox}{max totalsize={0.98\linewidth}{0.55\textheight},center}
\begin{minipage}{\linewidth}
\centering

\begin{tabular}{c}

\resizebox{0.98\linewidth}{!}{
\begin{tikzpicture}[
    font=\scriptsize,
    block/.style={
        draw,
        rounded corners,
        align=center,
        minimum height=0.82cm,
        minimum width=2.20cm,
        fill=blue!6
    },
    imgblock/.style={
        draw,
        rounded corners,
        align=center,
        inner sep=1.5pt,
        fill=gray!3
    },
    edgeblock/.style={
        draw=orange!80!black,
        rounded corners,
        align=center,
        minimum height=0.74cm,
        minimum width=2.05cm,
        fill=orange!10
    },
    mambablock/.style={
        draw=purple!80!black,
        line width=1.2pt,
        rounded corners,
        align=center,
        minimum height=0.95cm,
        minimum width=2.85cm,
        fill=purple!8
    },
    refineblock/.style={
        draw=green!50!black,
        rounded corners,
        align=center,
        minimum height=0.82cm,
        minimum width=2.25cm,
        fill=green!8
    },
    segblock/.style={
        draw=teal!70!black,
        rounded corners,
        align=center,
        minimum height=0.90cm,
        minimum width=2.65cm,
        fill=teal!8
    },
    stagebox/.style={
        draw,
        dashed,
        rounded corners,
        fill=gray!3,
        inner sep=0.22cm
    },
    arrow/.style={
        -{Latex[length=2.2mm]},
        thick
    },
    dashedarrow/.style={
        -{Latex[length=2.2mm]},
        thick,
        dashed
    }
]

\node[imgblock] (inputimg) at (0,0)
{
    \includegraphics[width=1.55cm]{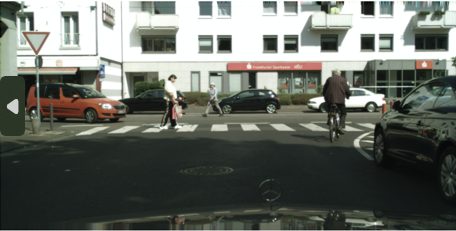}
};
\node[below=0.05cm of inputimg] {Input};

\node[block] (input) at (2.55,0)
{Input\\Image};

\node[block] (encoder) at (5.45,0)
{ConvNeXt-Tiny\\Encoder\\
{\tiny $F_1$--$F_4$}};

\node[edgeblock] (edge) at (5.45,-1.35)
{Sobel Edge\\Prior};

\node[block] (decoder) at (8.55,0)
{Multi-scale\\Decoder\\
{\tiny $D_2, D_3, D_4$}};

\node[mambablock] (mamba) at (12.20,0)
{Hierarchical\\Reload-Mamba\\
{\tiny ($l\in\{2,3,4\}$)}};

\node[refineblock] (refine) at (15.80,0)
{Detail Fuse\\
{\tiny + Edge Prior}};

\node[segblock] (seghead) at (19.30,0)
{Segmentation Head\\
{\tiny + Output Map}};

\node[imgblock] (outputimg) at (22.60,0)
{
    \includegraphics[width=1.55cm]{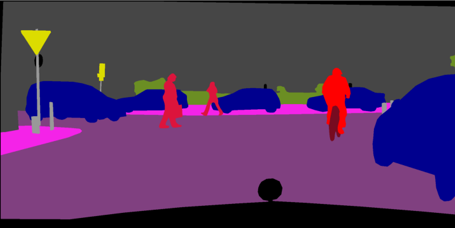}
};
\node[below=0.05cm of outputimg] {Output};

\draw[arrow] (inputimg.east) -- (input.west);
\draw[arrow] (input) -- (encoder);
\draw[arrow] (encoder) -- (decoder);
\draw[arrow] (decoder) -- (mamba);
\draw[arrow] (mamba) -- (refine);
\draw[arrow] (refine) -- (seghead);
\draw[arrow] (seghead.east) -- (outputimg.west);

\draw[arrow] (input.south) |- (edge.west);
\draw[dashedarrow] (edge.east) -| (decoder.south);

\coordinate (edgeout) at ($(edge.south)+(0.18,0)$);
\coordinate (edgeA) at ($(edgeout)+(0,-0.15)$);
\coordinate (refinein) at ($(refine.south)+(0,0)$);

\draw[dashedarrow]
(edgeout) -- (edgeA) -- (edgeA -| refinein) -- (refinein);

\begin{scope}[on background layer]
\node[
    stagebox,
    fit=(input)(encoder)(edge)(decoder),
    inner xsep=0.32cm,
    inner ysep=0.28cm
] (stage1) {};

\node[
    stagebox,
    draw=purple!70!black,
    fill=purple!2,
    fit=(mamba),
    inner xsep=0.34cm,
    inner ysep=0.28cm
] (stage2) {};

\node[
    stagebox,
    draw=green!45!black,
    fill=green!2,
    fit=(refine)(seghead),
    inner xsep=0.32cm,
    inner ysep=0.28cm
] (stage3) {};
\end{scope}

\node[
    font=\bfseries\scriptsize,
    above=0.12cm of stage1
]
{Stage 1: Feature Extraction};

\node[
    font=\bfseries\scriptsize,
    text=purple!80!black,
    above=0.12cm of stage2
]
{Stage 2: Hierarchical Context Modeling};

\node[
    font=\bfseries\scriptsize,
    text=green!50!black,
    above=0.12cm of stage3
]
{Stage 3: Refinement};

\end{tikzpicture}
}

\\[-0.08cm]

{\scriptsize\textcolor{purple!80!black}{Expanded view of the Hierarchical Reload-Mamba module (applied at $l \in \{2,3,4\}$)}}

\\[0.02cm]

\resizebox{0.96\linewidth}{!}{
\begin{tikzpicture}[
    font=\scriptsize,
    innerblock/.style={
        draw,
        rounded corners,
        align=center,
        minimum height=0.85cm,
        minimum width=2.45cm,
        fill=white
    },
    priorblock/.style={
        draw=blue!70!black,
        line width=1.0pt,
        rounded corners,
        align=center,
        minimum height=0.85cm,
        minimum width=2.95cm,
        fill=blue!10
    },
    gateblock/.style={
        draw=red!75!black,
        line width=1.15pt,
        rounded corners,
        align=center,
        minimum height=1.05cm,
        minimum width=3.55cm,
        fill=red!10
    },
    auxblock/.style={
        draw=orange!85!black,
        line width=1.0pt,
        rounded corners,
        align=center,
        minimum height=0.95cm,
        minimum width=3.10cm,
        fill=orange!12
    },
    bdsupblock/.style={
        draw=blue!70!black,
        rounded corners,
        align=center,
        minimum height=0.65cm,
        minimum width=2.40cm,
        fill=blue!6
    },
    deepsupblock/.style={
        draw=gray!60!black,
        rounded corners,
        align=center,
        minimum height=0.65cm,
        minimum width=2.10cm,
        fill=gray!8
    },
    hierblock/.style={
        draw=purple!75!black,
        line width=1.1pt,
        rounded corners,
        align=center,
        minimum height=1.0cm,
        minimum width=2.95cm,
        fill=purple!12
    },
    restoredblock/.style={
        draw,
        rounded corners,
        align=center,
        minimum height=0.85cm,
        minimum width=2.45cm,
        fill=white
    },
    purplebox/.style={
        draw=purple!75!black,
        dashed,
        rounded corners,
        fill=purple!2,
        inner sep=0.45cm
    },
    arrow/.style={
        -{Latex[length=2.2mm]},
        thick
    },
    dashedarrow/.style={
        -{Latex[length=2.2mm]},
        thick,
        dashed
    }
]

\node[bdsupblock] (bdsup) at (0, 1.7)
{GT Boundary $B_{gt}^{(l)}$};

\node[priorblock] (prior) at (0,0)
{Boundary-Supervised\\Detail Prior $P^{(l)}$};

\node[innerblock] (select) at (3.65,0)
{Detail\\Selection $X_d^{(l)}$};

\node[innerblock] (scan) at (7.20,0)
{4-Directional\\Scan};

\node[innerblock] (fusion) at (10.75,0)
{Directional\\Fusion $M^{(l)}$};

\node[gateblock] (reload) at (14.75,0)
{Class-Uncertainty-Aware\\Reload Gate};

\node[restoredblock] (restored) at (18.65,0)
{Restored\\$D_m^{(l)}$};

\node[auxblock] (aux) at (10.75, -2.1)
{Pre-reload Aux Head\\$Y_u^{(l)} \to U^{(l)}$};

\node[deepsupblock] (deepsup) at (6.20, -2.1)
{Deep\\Supervision};

\node[hierblock] (hierfuse) at (18.65, -2.1)
{Top-Down\\Hierarchical Fusion\\$\to D_m$};

\draw[arrow] (prior) -- (select);
\draw[arrow] (select) -- (scan);
\draw[arrow] (scan) -- (fusion);
\draw[arrow] (fusion) -- (reload);
\draw[arrow] (reload) -- (restored);

\draw[dashedarrow, draw=blue!70!black, line width=0.8pt]
    (bdsup.south) -- (prior.north);
\node[font=\tiny, color=blue!70!black, anchor=west]
    at ($(bdsup.south)!0.5!(prior.north) + (0.08,0)$) {$\mathcal{L}_{prior}$};

\draw[arrow, draw=orange!85!black, line width=0.85pt]
    (fusion.south) -- (aux.north);

\draw[arrow, draw=orange!85!black, line width=0.85pt]
    (aux.east) -| (reload.south);
\node[font=\tiny, color=orange!85!black]
    at ($(aux.east) + (1.6, 0.25)$) {Uncertainty $U^{(l)}$};

\draw[dashedarrow, draw=gray!60!black, line width=0.75pt]
    (aux.west) -- (deepsup.east);

\draw[arrow, draw=purple!75!black, line width=0.95pt]
    (restored.south) -- (hierfuse.north);
\node[font=\tiny, color=purple!75!black, anchor=west]
    at ($(restored.south)!0.45!(hierfuse.north) + (0.08,0)$) {for $l \in \{2,3,4\}$};

\begin{scope}[on background layer]
\node[
    purplebox,
    fit=(prior)(select)(scan)(fusion)(reload)(restored)(aux)(deepsup)(hierfuse)(bdsup),
    inner xsep=0.55cm,
    inner ysep=0.45cm
] {};
\end{scope}

\end{tikzpicture}
}

\end{tabular}

\end{minipage}
\end{adjustbox}

\vspace{0.05cm}


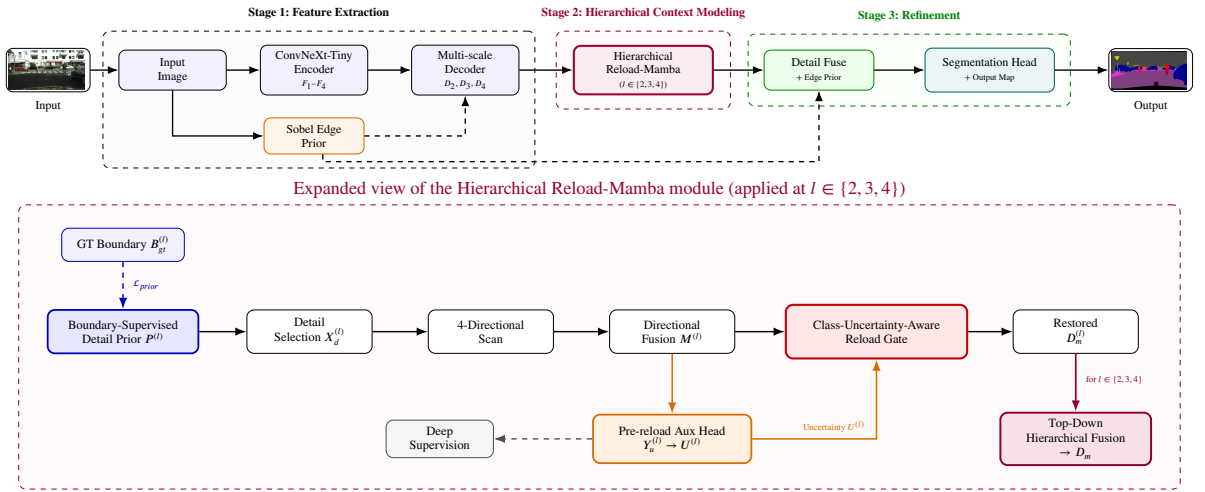
\captionof{figure}{Overall architecture of the proposed Reload-Mamba segmentation network. The upper part shows the main pipeline. The lower part expands the Hierarchical Reload-Mamba module, which is applied at three decoder levels $l \in \{2, 3, 4\}$ and integrates three segmentation-specific designs: (i) a \textbf{boundary-supervised detail prior} (blue) trained with the ground-truth boundary mask $B_{gt}^{(l)}$ via $\mathcal{L}_{prior}$; (ii) a \textbf{class-uncertainty-aware Reload Gate} (red) that takes a per-pixel entropy map $U^{(l)}$ produced by a pre-reload auxiliary head (orange) as an additional input; and (iii) a \textbf{top-down hierarchical fusion} (purple) that aggregates the restored features $\{D_m^{(2)}, D_m^{(3)}, D_m^{(4)}\}$ across decoder levels to produce the final $D_m$.}
\label{fig:reload_mamba_arch}

\end{minipage}

\medskip

\subsection{Encoder-Decoder Backbone}
The backbone of Reload-Mamba is built upon ConvNeXt-Tiny \cite{liu2022convnext}, which provides hierarchical feature maps at four different scales. Let the encoder outputs be denoted as
\begin{equation}
    \{F_1, F_2, F_3, F_4\} = E(I),
\end{equation}
where $F_1$, $F_2$, $F_3$, and $F_4$ correspond to feature maps with channel dimensions 96, 192, 384, and 768, respectively. These multi-scale features contain both low-level spatial information and high-level semantic information. The decoder follows a U-shaped skip-fusion design. The deepest feature $F_4$ is first upsampled and concatenated with $F_3$, and the process is then repeated with $F_2$ and $F_1$:
\begin{equation}
    D_4 = \phi_4([\mathrm{Up}(F_4), F_3]),
\end{equation}
\begin{equation}
    D_3 = \phi_3([\mathrm{Up}(D_4), F_2]),
\end{equation}
\begin{equation}
    D_2 = \phi_2([\mathrm{Up}(D_3), F_1]),
\end{equation}
where $[\cdot,\cdot]$ denotes channel-wise concatenation, $\mathrm{Up}(\cdot)$ denotes transposed convolution-based upsampling, and $\phi_i(\cdot)$ denotes a convolutional refinement block. Unlike single-level Mamba refinement designs that operate only on the finest decoder feature, the three intermediate decoder features $\{D_2, D_3, D_4\}$ are all forwarded to the proposed hierarchical Reload-Mamba module (Section~\ref{sec:multi_level}), so that anti-dilution refinement can be applied at three semantically distinct scales.

\subsection{Sobel Edge Prior Branch}
Since semantic segmentation requires accurate object boundaries, Reload-Mamba introduces an explicit Sobel edge prior branch. Given the input image $I$, horizontal and vertical gradient responses are extracted using Sobel operators:
\begin{equation}
    G_x = \mathrm{Sobel}_x(I), \quad
    G_y = \mathrm{Sobel}_y(I).
\end{equation}
The original image and the absolute gradient responses are concatenated:
\begin{equation}
    E_s = \psi_s([I, |G_x|, |G_y|]),
\end{equation}
where $\psi_s(\cdot)$ denotes a convolutional projection module for transforming edge responses into boundary-aware features. The output edge prior $E_s \in \mathbb{R}^{32 \times H \times W}$ provides low-level boundary cues for the final detail fusion stage, complementing the learned boundary-supervised prior introduced in the next subsection.

\subsection{Boundary-Supervised Local Detail Prior and Directional Scanning}
\label{sec:prior_and_scan}
In our prior single-level anti-dilution architecture~\cite{chan2026deepmine}, the local detail prior is estimated implicitly from the decoder feature, relying on the downstream task loss to shape its spatial pattern. This implicit estimation is adequate for binary stroke restoration, where foreground pixels are by definition the regions to preserve. In multi-class semantic segmentation, however, the regions that need response restoration---class boundaries, thin structures, and small categories---are not uniquely identifiable from class probabilities alone. We therefore explicitly supervise the local detail prior using ground-truth boundary masks, providing a localized and stable conditioning signal independently of the downstream task loss landscape.

Given a decoder feature $D_l$ at level $l \in \{2, 3, 4\}$, the level-specific prior map is generated as
\begin{equation}
    P^{(l)} = \sigma(\psi_p^{(l)}(D_l)),
    \label{eq:prior_level}
\end{equation}
where $\psi_p^{(l)}(\cdot)$ is a level-specific convolutional projection and $\sigma(\cdot)$ is the sigmoid function. A binary boundary mask $B_{gt}$ is derived from the semantic label map $G$, where $B_{gt,i}=1$ if the $3 \times 3$ neighborhood centered at pixel $i$ contains different semantic labels, and $0$ otherwise. The boundary mask is downsampled to the resolution of $P^{(l)}$ via max-pooling to preserve boundary signal under spatial downsampling, producing $B_{gt}^{(l)}$. Each prior is then directly supervised by a binary cross-entropy loss:
\begin{equation}
    \mathcal{L}_{prior}^{(l)} = -\frac{1}{|\Omega_l|}\sum_{i \in \Omega_l}\!\left[B_{gt,i}^{(l)}\log P_i^{(l)} + (1-B_{gt,i}^{(l)})\log(1-P_i^{(l)})\right]\!,
    \label{eq:prior_loss}
\end{equation}
where $\Omega_l$ is the set of valid pixels at level $l$. The total prior loss aggregates across levels:
\begin{equation}
    \mathcal{L}_{prior} = \frac{1}{3}\sum_{l \in \{2,3,4\}} \mathcal{L}_{prior}^{(l)}.
    \label{eq:prior_loss_total}
\end{equation}
This explicit supervision ensures that $P^{(l)}$ reliably highlights boundary-sensitive and detail-sensitive regions, providing a stable conditioning signal for the downstream Reload Gate.

Given the supervised prior, a detail-aware feature is constructed at each level as
\begin{equation}
    X_d^{(l)} = D_l \odot P^{(l)} + \lambda_{low}\, D_l \odot (1-P^{(l)}),
\end{equation}
where $\odot$ denotes element-wise multiplication and $\lambda_{low}$ retains a controlled amount of low-priority contextual responses. We set $\lambda_{low}=0.3$ in our implementation.

After detail-aware feature selection, $X_d^{(l)}$ is processed by four directional Mamba scans. Each directional scan $\mathcal{M}_{*}^{(l)}$ is implemented as a selective state-space block following the Mamba formulation \cite{gu2023mamba}. The feature map $X_d^{(l)}$ is flattened into a 1D token sequence according to the corresponding traversal order (rightward, leftward, downward, or upward), processed by a Mamba block with input-dependent SSM parameters $(A, B, C, \Delta)$, and reshaped back to 2D. The four directional scans do not share parameters. Different from the 2D selective scan of VMamba \cite{liu2024vmamba}, which interleaves multiple traversals into a unified token sequence inside the SSM kernel, Reload-Mamba treats each direction as a separate SSM pass and aggregates the outputs through pixel-wise directional attention. Specifically,
\begin{equation}
    H_r^{(l)} = \mathcal{M}_{r}^{(l)}(X_d^{(l)}), \quad
    H_l^{(l)} = \mathcal{M}_{l}^{(l)}(X_d^{(l)}),
\end{equation}
\begin{equation}
    H_d^{(l)} = \mathcal{M}_{d}^{(l)}(X_d^{(l)}), \quad
    H_u^{(l)} = \mathcal{M}_{u}^{(l)}(X_d^{(l)}).
\end{equation}
The four directional features are then fused by pixel-wise directional attention:
\begin{equation}
    A^{(l)} = \mathrm{Softmax}(\psi_a^{(l)}([H_u^{(l)}, H_d^{(l)}, H_l^{(l)}, H_r^{(l)}])),
    \label{eq:dir_attn}
\end{equation}
\begin{equation}
    M^{(l)} = A_u^{(l)} \odot H_u^{(l)} + A_d^{(l)} \odot H_d^{(l)} + A_l^{(l)} \odot H_l^{(l)} + A_r^{(l)} \odot H_r^{(l)}.
    \label{eq:dir_fuse}
\end{equation}
This design enables each pixel to adaptively select useful directional information before the Reload Gate restores detail-sensitive local responses.

\subsection{Class-Uncertainty-Aware Reload Gate}
\label{sec:reload_gate}
Sequential state-space propagation may attenuate local responses, especially around object boundaries, thin structures, and small semantic regions. To address this issue, we introduce a class-uncertainty-aware Reload Gate. In our prior binarization architecture~\cite{chan2026deepmine}, the gate conditions only on the original feature and the propagation-induced feature change $|D_l - M^{(l)}|$, treating each pixel uniformly regardless of its semantic ambiguity; this is sufficient when ambiguity is one-dimensional, as in binary foreground restoration. In multi-class segmentation, however, the regions most affected by propagation-induced dilution typically coincide with high per-pixel class entropy. We therefore additionally exploit per-pixel class uncertainty as a third gating input, a signal that is meaningful only when the number of classes exceeds two.

At each decoder level $l$, we first attach a lightweight pre-reload auxiliary head $\psi_{u}^{(l)}(\cdot)$ to the directional fusion output $M^{(l)}$ to produce preliminary semantic logits:
\begin{equation}
    Y_u^{(l)} = \psi_{u}^{(l)}(M^{(l)}), \qquad
    Y_u^{(l)} \in \mathbb{R}^{C \times H_l \times W_l}.
    \label{eq:pre_aux}
\end{equation}
This head is computed before the Reload Gate so that it does not introduce a circular dependency. The corresponding per-pixel normalized entropy is
\begin{equation}
    U_i^{(l)} = -\frac{1}{\log C}\sum_{c=1}^{C} P_{u,c,i}^{(l)}\log P_{u,c,i}^{(l)},
    \qquad
    P_u^{(l)} = \mathrm{Softmax}(Y_u^{(l)}),
    \label{eq:uncertainty}
\end{equation}
where $C$ is the number of semantic classes and the normalization by $\log C$ ensures $U_i^{(l)} \in [0,1]$. Higher $U_i^{(l)}$ indicates greater per-pixel classification ambiguity, which empirically coincides with the regions most affected by propagation-induced dilution.

The Reload Gate then jointly considers the original decoder feature, the propagation-induced feature change, and the class uncertainty:
\begin{equation}
    I_d^{(l)} = \sigma(\psi_d^{(l)}([\,D_l,\; |D_l - M^{(l)}|,\; U^{(l)}\,])),
    \label{eq:gate_new}
\end{equation}
where $\psi_d^{(l)}(\cdot)$ is a lightweight convolutional detector. This formulation is fundamentally specific to multi-class segmentation: in binary tasks, the entropy term collapses to a foreground--background uncertainty already captured by the prior, so the additional input becomes informative only when $C > 2$.

The restoration map is further modulated by the boundary-supervised prior:
\begin{equation}
    \hat{I}_d^{(l)} = \alpha\, P^{(l)} \odot I_d^{(l)},
\end{equation}
where $\alpha$ controls the maximum restoration strength. We set $\alpha=0.7$. The level-specific restored feature is computed as
\begin{equation}
    D_m^{(l)} = M^{(l)} + \hat{I}_d^{(l)} \odot (D_l - M^{(l)}).
    \label{eq:reload}
\end{equation}
This bounded reloading step restores information from $D_l$ when a location is both detail-sensitive ($P^{(l)}$ large), shows large propagation-induced change ($|D_l - M^{(l)}|$ large), and is semantically ambiguous ($U^{(l)}$ large). Otherwise, the model relies more on the Mamba-enhanced contextual representation $M^{(l)}$.

\subsection{Hierarchical Multi-Level Reload}
\label{sec:multi_level}
The Reload Gate described above is applied independently at each decoder level $l \in \{2, 3, 4\}$ using level-specific parameters. This hierarchical design extends single-level anti-dilution refinement, which applies the gate at a single feature resolution, and is motivated by the observation that propagation-induced dilution affects different semantic scales differently: large coherent regions are best refined at coarse resolutions, medium objects at $D_3$, and thin structures and boundaries at the finest level $D_2$.

To control the parameter overhead introduced by additional Mamba blocks at $D_3$ and $D_4$, we use level-dependent channel widths: $D_4$ uses a reduced width of $c_4$, $D_3$ uses $c_3$, and $D_2$ uses the original $c_2=96$. In our implementation, we set $(c_4, c_3, c_2) = (192, 128, 96)$. Section~\ref{sec:ablation_channels} ablates this choice.

The level-specific restored features $\{D_m^{(2)}, D_m^{(3)}, D_m^{(4)}\}$ are progressively fused in a top-down manner, mirroring the decoder structure:
\begin{equation}
    \tilde{D}_3 = \phi_3^h\!\left(\left[\mathrm{Up}(D_m^{(4)}),\; D_m^{(3)}\right]\right),
\end{equation}
\begin{equation}
    D_m = \phi_2^h\!\left(\left[\mathrm{Up}(\tilde{D}_3),\; D_m^{(2)}\right]\right),
    \label{eq:hier_fuse}
\end{equation}
where $\phi_l^h(\cdot)$ are lightweight depthwise-separable convolutional fusion blocks, and $D_m \in \mathbb{R}^{96 \times H/4 \times W/4}$ is the hierarchical Reload-Mamba output forwarded to the downstream detail fusion stage. This cross-level fusion enforces consistency between restored responses at different resolutions, which is particularly useful for stabilizing predictions of small semantic categories.

\subsection{Relation to Our Prior Anti-Dilution Architecture}
\label{sec:relation}
Reload-Mamba extends the anti-dilution principle introduced in our prior work on Mamba-based document image binarization~\cite{chan2026deepmine}, which uses (i) an implicitly estimated local prior, (ii) a gate conditioned only on the original feature and the propagation-induced feature change, and (iii) refinement at a single decoder resolution. While that single-level design is effective for binary stroke restoration, multi-class semantic segmentation presents qualitatively different requirements that motivate the three segmentation-specific designs introduced in Sections~\ref{sec:prior_and_scan}--\ref{sec:multi_level}. We summarize the technical differences below.
\begin{itemize}
    \item \textbf{Boundary-supervised vs.\ implicit prior.} Our prior architecture estimates the prior implicitly from the decoder feature, shaping it through the downstream binarization loss alone. In multi-class segmentation, boundary regions are not uniquely identifiable from class probabilities, so we instead supervise the prior explicitly using ground-truth boundary masks (Eq.~(\ref{eq:prior_loss})). The empirical impact of this choice is reported in Section~\ref{sec:ablation_prior_sup}.
    \item \textbf{Class-uncertainty-aware vs.\ uncertainty-agnostic gate.} The gate in our prior architecture conditions only on the feature change $|D_l - M^{(l)}|$, which is sufficient when ambiguity is one-dimensional. In multi-class segmentation, regions that benefit from response restoration coincide with high per-pixel class entropy. We therefore incorporate normalized entropy of the pre-reload auxiliary logits into the gate input (Eq.~(\ref{eq:gate_new})). This term is informative only when $C > 2$ and is therefore fundamentally specific to multi-class dense prediction.
    \item \textbf{Hierarchical multi-level vs.\ single-level Reload.} Our prior architecture applies the anti-dilution gate at one decoder resolution. Reload-Mamba instead applies the gate at three decoder levels $\{D_2, D_3, D_4\}$ and fuses the restored responses top-down (Eq.~(\ref{eq:hier_fuse})), addressing propagation-induced dilution across distinct semantic scales.
\end{itemize}
The shared design ingredient between Reload-Mamba and our prior binarization architecture is the use of a comparison between pre-propagation and post-propagation features to estimate response attenuation. Beyond this shared ingredient, the prior, the gate input, and the placement of refinement are all redesigned for multi-class segmentation, and the cumulative effect of these design choices is verified in Section~\ref{sec:ablation_cumulative}.

\subsection{Edge Guided Detail Fusion and Segmentation Head}
The hierarchical Reload-Mamba output $D_m$ is upsampled back to the original spatial resolution:
\begin{equation}
    D_1 = \mathrm{Up}_{4\times}(D_m).
\end{equation}
The upsampled feature is concatenated with the Sobel edge prior $E_s$ defined in Section~3.2:
\begin{equation}
    D_c = [D_1,\; E_s].
\end{equation}
A detail fusion module then produces a boundary-aware feature:
\begin{equation}
    D_f = \psi_f(D_c),
\end{equation}
where $\psi_f(\cdot)$ consists of convolutional and depthwise convolutional operations. Before final prediction, a gate is used to suppress noisy activations:
\begin{equation}
    P_g = \sigma(\psi_g(D_f)),
\end{equation}
\begin{equation}
    \hat{D}_f = D_f \odot (0.5 + 0.5\, P_g).
\end{equation}
The final segmentation logits are then produced by the segmentation head:
\begin{equation}
    Y = \psi_{seg}(\hat{D}_f),
\end{equation}
where $Y \in \mathbb{R}^{C \times H \times W}$ and $C$ is the number of semantic classes (e.g., $C=19$ for Cityscapes). The pre-reload auxiliary logits $\{Y_u^{(l)}\}_{l \in \{2,3,4\}}$ from Eq.~(\ref{eq:pre_aux}) are upsampled to the original resolution and used as auxiliary predictions for deep supervision during training; only the final segmentation output $Y$ is used at inference time.

\subsection{Training Objective}
Reload-Mamba is trained using a composite segmentation objective that combines hard-pixel cross-entropy, region-level IoU optimization, boundary-aware supervision, multi-level auxiliary deep supervision, and the prior supervision introduced in Section~\ref{sec:prior_and_scan}. Pixels with the ignore label are excluded from all loss computations. The total loss is defined as
\begin{equation}
    \mathcal{L}_{total}
    =
    \mathcal{L}_{ohem}
    + \lambda_{lovasz}\mathcal{L}_{lovasz}
    + \lambda_{bd}\mathcal{L}_{bd}
    + \lambda_{aux}\mathcal{L}_{aux}
    + \lambda_{prior}\mathcal{L}_{prior}.
    \label{eq:total_loss}
\end{equation}
We set $\lambda_{lovasz}=0.8$, $\lambda_{bd}=0.1$, $\lambda_{aux}=0.4$, and $\lambda_{prior}=0.2$.

For pixel-wise semantic supervision, we first compute the cross-entropy loss for each valid pixel:
\begin{equation}
    \ell_i = -\log\frac{\exp(Y_{G_i,i})}{\sum_{c=1}^{C}\exp(Y_{c,i})},
    \quad i \in \Omega,
\end{equation}
where $\Omega$ denotes the set of valid pixels. Online hard example mining is then applied: the valid pixel losses are sorted in descending order, and the top $\rho |\Omega|$ pixels are retained, with $\rho=0.85$ in our implementation. The OHEM loss is
\begin{equation}
    \mathcal{L}_{ohem} = \frac{1}{|\Omega_h|}\sum_{i \in \Omega_h}\ell_i,
\end{equation}
where $\Omega_h$ is the selected hard-pixel set.

To better align training with mIoU, we additionally use the Lovász-Softmax loss:
\begin{equation}
    \mathcal{L}_{lovasz}
    =
    \frac{1}{|\mathcal{C}_{p}|}\sum_{c \in \mathcal{C}_{p}}\overline{\Delta_{\mathrm{Jaccard}}}(m(c)),
\end{equation}
where $\mathcal{C}_{p}$ denotes the set of classes present in the current batch, $m(c)$ denotes the sorted prediction errors of class $c$, and $\overline{\Delta_{\mathrm{Jaccard}}}$ is the Lovász extension of the Jaccard loss.

The boundary-weighted cross-entropy term emphasizes pixels near semantic transitions. The boundary mask $B$ is generated from the ground-truth label map as in Eq.~(\ref{eq:prior_loss}) (without downsampling, at the full output resolution):
\begin{equation}
    B_i =
    \begin{cases}
    1, & \max_{j \in \mathcal{N}_3(i)} G_j \neq \min_{j \in \mathcal{N}_3(i)} G_j,\\
    0, & \text{otherwise}.
    \end{cases}
\end{equation}
The boundary loss is then
\begin{equation}
    \mathcal{L}_{bd} = \frac{\sum_{i \in \Omega} B_i \ell_i}{\sum_{i \in \Omega} B_i + \epsilon}.
\end{equation}

The multi-level auxiliary loss aggregates standard cross-entropy supervision on the three pre-reload auxiliary heads $\{Y_u^{(l)}\}$:
\begin{equation}
    \mathcal{L}_{aux}
    =
    \frac{1}{3}\sum_{l \in \{2,3,4\}}
    \frac{1}{|\Omega_l|}\sum_{i \in \Omega_l}\mathrm{CE}(Y_{u,i}^{(l)}, G_i^{(l)}),
\end{equation}
where $G^{(l)}$ is the ground-truth label downsampled to the resolution of $Y_u^{(l)}$ via nearest-neighbor interpolation. This multi-level supervision stabilizes both the directional fusion outputs and the uncertainty estimates used inside the Reload Gate. The prior loss $\mathcal{L}_{prior}$ is defined in Eq.~(\ref{eq:prior_loss_total}).

All auxiliary losses are used only during training; at inference, only the final segmentation output $Y$ is produced.

\section{Experiments}
\subsection{Datasets}
\textbf{Cityscapes.}
Cityscapes is a widely used urban scene understanding benchmark for semantic segmentation \cite{cordts2016cityscapes}. It contains high-resolution street-view images with fine pixel-level annotations. Following the standard setting, we evaluate the model on 19 semantic classes. The dataset is challenging because it contains complex road scenes, thin objects, traffic signs, poles, pedestrians, and boundary-sensitive regions. Therefore, it is suitable for evaluating whether Reload-Mamba can preserve local details while modeling global context.

\textbf{ADE20K.}
ADE20K is a large-scale scene parsing benchmark containing diverse indoor and outdoor scenes \cite{zhou2017ade20k}. Compared with Cityscapes, ADE20K contains more semantic categories and more complex object layouts. We use ADE20K to evaluate the generalization ability of Reload-Mamba under multi-class semantic segmentation. The dataset is especially useful for testing whether the proposed Reload Gate can remain effective when the segmentation task involves diverse object scales and scene structures.

\textbf{PASCAL VOC 2012.}
PASCAL VOC is a commonly used semantic segmentation benchmark \cite{everingham2010pascal}. Following common practice, we use the augmented training set composed of PASCAL VOC and the Semantic Boundaries Dataset (SBD) \cite{hariharan2011sbd}. Compared with Cityscapes and ADE20K, PASCAL VOC contains more object-centric images and provides a complementary evaluation setting for validating the generality of Reload-Mamba.

\subsection{Implementation Details}
All experiments are implemented using PyTorch \cite{paszke2019pytorch} and conducted on a single NVIDIA RTX 5080 GPU. Reload-Mamba is evaluated under two backbone settings. For ADE20K and Cityscapes, we adopt ConvNeXt-Tiny as the encoder backbone \cite{liu2022convnext}, initialized with ImageNet-1K pretrained weights \cite{deng2009imagenet}. For PASCAL VOC, we adopt ResNet-101 with output stride 8 as the encoder backbone, following the common DeepLab-style PASCAL VOC training protocol \cite{chen2017deeplab,chen2018deeplabv3plus}. Specifically, the encoder is initialized with MS-COCO \cite{lin2014coco} pre-trained weights and then fine-tuned on the augmented PASCAL VOC training set composed of PASCAL VOC and SBD. This is consistent with the standard practice adopted by representative ResNet-101 based VOC segmentation methods such as DeepLabv3, EMANet~\cite{li2019emanet}, CFNet~\cite{zhang2019cfnet}, and DeepLabv3+. For ADE20K and Cityscapes, no COCO pre-training is used; only ImageNet-1K \cite{deng2009imagenet} pre-trained weights are loaded into the encoder. Mapillary Vistas \cite{neuhold2017mapillary} is not used in any of our experiments. This allows us to evaluate the contribution of the proposed Reload Gate and directional state-space modeling modules without relying on extra segmentation pre-training. The model is optimized using the AdamW optimizer \cite{loshchilov2019adamw} with an initial learning rate of $1 \times 10^{-4}$ and a polynomial learning rate decay schedule with a power of 0.9. All models are trained for 200 epochs. The mini-batch size is set to 4. To stabilize optimization under single-GPU memory constraints, we use gradient accumulation with an accumulation step of 4. Therefore, the effective batch size is 16, while the forward pass operations are still computed with a mini-batch size of 4. For Cityscapes, the crop size is set to $768 \times 768$. For ADE20K, the crop size is set to $512 \times 512$. For PASCAL VOC, we train on the augmented training set composed of PASCAL VOC and the Semantic Boundaries Dataset (SBD), and the crop size is also set to $512 \times 512$. The number of semantic classes is set according to each dataset. Specifically, Cityscapes contains 19 classes, ADE20K contains 150 classes, and PASCAL VOC contains 21 classes including background. During training, we apply common data augmentation strategies, including random scaling with a ratio between 0.5 and 2.0, random horizontal flipping, and random color jittering. Unless otherwise specified, all quantitative results are reported under single-scale inference without test-time augmentation. For multi-scale testing, we use scaling factors of $\{0.5, 0.75, 1.0, 1.25, 1.5\}$ with horizontal flipping.

\subsection{Comparison with Existing Methods}
We compare Reload-Mamba with representative semantic segmentation methods on ADE20K, Cityscapes, and PASCAL VOC. Since the available comparison protocols are not symmetric across datasets, we organize the comparisons according to the reliability and commonality of publicly reported results. ADE20K is used as the primary benchmark because ConvNeXt-Tiny, SegFormer, SegNeXt, and several Mamba-based models are commonly evaluated under comparable settings. Cityscapes is used as a secondary benchmark, where the results are more sensitive to crop size, training split, and single-scale or multi-scale inference. PASCAL VOC is reported as an additional benchmark because modern Transformer-based and Mamba-based segmentation models rarely report VOC results. For ADE20K and Cityscapes, Reload-Mamba uses a ConvNeXt-Tiny encoder. For PASCAL VOC, Reload-Mamba uses ResNet-101 with output stride 8 to follow the common ResNet-based VOC comparison setting. Therefore, VOC results should be interpreted as a complementary evaluation under a ResNet-101 segmentation setting, rather than a direct comparison with the ConvNeXt-Tiny results on ADE20K and Cityscapes. It is worth noting that parameter counts are not always directly comparable across different segmentation frameworks. For example, ADE20K results based on UPerNet usually report the parameters of both the backbone and the UPerNet decoder head, while SegFormer and SegNeXt often use lighter segmentation heads. Therefore, we focus primarily on models of similar practical scale and evaluation settings rather than comparing backbone-only parameter counts.

\subsubsection{Results on ADE20K}

ADE20K is the main comparison benchmark in this work. Table~\ref{tab:comparison_ade20k} reports the validation mIoU of representative compact and mid-scale segmentation models. Reload-Mamba is compared with CNN-based, Transformer-based, and Mamba-based methods under commonly reported single-scale and multi-scale settings. In particular, ConvNeXt-T, Swin-T, VMamba-T, MambaOut-T, and MambaVision-T provide meaningful references because they are commonly evaluated under comparable ADE20K segmentation settings.

\begin{table*}[htbp]
\centering
\caption{Comparison with existing methods on the ADE20K validation set. SS and MS denote single-scale and multi-scale inference, respectively. Parameter counts may include both backbone and segmentation head depending on the original implementation.}
\label{tab:comparison_ade20k}
\begin{tabular}{lccc}
\hline
Method & Backbone & Params & mIoU SS / MS (\%) \\
\hline
SegFormer-B0 \cite{xie2021segformer} & MiT-B0 & 3.8M & 37.4 / 38.0 \\
SegNeXt-T \cite{guo2022segnext} & MSCAN-T & 4.3M & 41.1 / 42.2 \\
SegFormer-B1 \cite{xie2021segformer} & MiT-B1 & 13.7M & 42.2 / 43.1 \\
SegNeXt-S \cite{guo2022segnext} & MSCAN-S & 13.9M & 44.3 / 45.8 \\
SegFormer-B2 \cite{xie2021segformer} & MiT-B2 & 27.5M & 46.5 / 47.5 \\
SegNeXt-B \cite{guo2022segnext} & MSCAN-B & 27.6M & 48.5 / 49.9 \\
\hline
Swin-T \cite{liu2021swin} & Swin-T & 60M & 44.4 / 45.8 \\
ConvNeXt-T \cite{liu2022convnext} & ConvNeXt-T & 60M & 46.0 / 46.7 \\
Focal-T \cite{yang2021focal} & Focal-T & 62M & 45.8 / -- \\
\hline
Vim-T \cite{zhu2024vim} & Vim-T & 13M & 41.0 / -- \\
Vim-S \cite{zhu2024vim} & Vim-S & 46M & 44.9 / -- \\
LocalMamba-T \cite{huang2024localvim} & LocalVim-T & 36M & 43.4 / 44.4 \\
MambaOut-T \cite{yu2024mambaout} & MambaOut-T & 54M & 47.4 / 48.6 \\
MambaVision-T \cite{hatamizadeh2024mambavision} & MambaVision-T & 55M & 46.0 / -- \\
VMamba-T \cite{liu2024vmamba} & VMamba-T & 55--62M & 47.3--47.9 / 48.8 \\
Spatial-Mamba-T \cite{xiao2024spatialmamba} & Spatial-Mamba-T & 57M & 48.6 / 49.4 \\
Spatial-Mamba-S \cite{xiao2024spatialmamba} & Spatial-Mamba-S & 73M & 50.6 / 51.4 \\
SparX-Mamba-S \cite{lou2024sparx} & SparX-Mamba-S & 77M & 51.3 / 52.5 \\
SegMAN-S \cite{fu2025segman} & SegMAN Encoder-S & 29.4M & 51.3 / -- \\
SegMAN-B \cite{fu2025segman} & SegMAN Encoder-B & 51.8M & 52.6 / -- \\
\hline
Reload-Mamba & ConvNeXt-Tiny + multi-scale decoder & 51M & \textbf{47.9} / \textbf{48.9} \\
\hline
\end{tabular}
\end{table*}

Reload-Mamba achieves 47.9\% single-scale and 48.9\% multi-scale mIoU with a ConvNeXt-Tiny backbone and 51M parameters. Compared with the ConvNeXt-T segmentation baseline (46.0\% SS, 60M params), Reload-Mamba demonstrates that adding directional state-space propagation together with hierarchical anti-dilution refinement improves dense prediction by $+1.9$ mIoU while using fewer parameters. Compared with VMamba-T (47.3--47.9\% SS, 55--62M params), Reload-Mamba matches or slightly outperforms the upper end of the VMamba-T range while being noticeably smaller; this contrast suggests that explicitly addressing response dilution through hierarchical Reload Gates is complementary to architectural improvements in selective scanning. Compared with SegFormer-B0/B1/B2, Reload-Mamba provides a different modeling strategy by replacing attention-based long-range modeling with directional state-space propagation and confirms the competitive accuracy--efficiency trade-off of Mamba-based segmentation models. Compared with the compact MambaOut-T (47.4\%) and Mamba-based variants Vim-T/S and LocalMamba-T, Reload-Mamba explicitly focuses on the response dilution problem caused by sequential propagation and restores local details through a class-uncertainty-aware Reload Gate. We further include recent Mamba-based segmentation networks proposed in 2024--2025, namely Spatial-Mamba~\cite{xiao2024spatialmamba}, SparX-Mamba~\cite{lou2024sparx}, and SegMAN~\cite{fu2025segman}, in Table~\ref{tab:comparison_ade20k} for completeness. These methods explore design directions that are distinct from ours: Spatial-Mamba introduces structure-aware state fusion inside the backbone, SparX-Mamba designs a sparse cross-layer connection mechanism between backbone stages, and SegMAN integrates sliding local attention with state-space modules inside the encoder. As shown in the table, these backbone-level redesigns attain higher mIoU at comparable parameter budgets, with SegMAN-B reaching 52.6\% at 51.8M parameters. We position our contribution as orthogonal to such backbone-level advances: rather than redesigning the visual state-space backbone itself, Reload-Mamba explicitly targets the propagation-induced response dilution that arises whenever sequential state-space scanning is applied to dense prediction, and instantiates this idea on top of a standard ConvNeXt-Tiny encoder with a multi-scale decoder. The proposed hierarchical Reload mechanism is therefore in principle compatible with these stronger Mamba-based backbones, and a systematic study of such integrations is left to future work. Accordingly, our ADE20K results should be interpreted as a controlled study of segmentation-specific anti-dilution refinement on a standard pipeline rather than as a SOTA-chasing comparison at the backbone level.

\subsubsection{Results on Cityscapes}

Table~\ref{tab:comparison_cityscapes} reports the comparison on Cityscapes. Unlike ADE20K, Cityscapes does not provide a widely adopted official ConvNeXt-Tiny segmentation baseline. Moreover, Cityscapes results are highly sensitive to crop size, training split, and whether multi-scale testing or additional data is used. Therefore, we compare Reload-Mamba mainly with representative compact Transformer-based methods, CNN-based segmentation methods, and available Mamba-based results, while keeping the differences in training protocols explicit.

\begin{table*}[htbp]
\centering
\caption{Comparison with existing methods on the Cityscapes validation set. Results may differ due to crop size, training split, additional data, and inference protocol. Reload-Mamba is evaluated under single-scale inference on the validation set.}
\label{tab:comparison_cityscapes}
\begin{tabular}{lccc}
\hline
Method & Backbone & Setting & mIoU (\%) \\
\hline
SegFormer-B0 \cite{xie2021segformer} & MiT-B0 & SS & 76.2 \\
SegFormer-B1 \cite{xie2021segformer} & MiT-B1 & SS & 78.5 \\
SegFormer-B2 \cite{xie2021segformer} & MiT-B2 & SS & 81.0 \\
SegNeXt-T \cite{guo2022segnext} & MSCAN-T & SS & 79.8 \\
SegNeXt-S \cite{guo2022segnext} & MSCAN-S & SS & 81.3 \\
SegNeXt-B \cite{guo2022segnext} & MSCAN-B & SS & 82.6 \\
SegMAN-T \cite{fu2025segman} & SegMAN Encoder-T & SS & 80.3 \\
SegMAN-B \cite{fu2025segman} & SegMAN Encoder-B & SS & 83.8 \\
\hline
PSPNet \cite{zhao2017pspnet} & ResNet-101 & val & 78.5 \\
DeepLabv3 \cite{chen2017deeplab} & ResNet-101 & val & 79.3 \\
CCNet \cite{huang2019ccnet} & ResNet-101 & val & 81.3 \\
OCRNet \cite{yuan2020ocrnet} & ResNet-101 & val & 81.8 \\
AlignSeg \cite{huang2020alignseg} & ResNet-101 & val & 82.4 \\
\hline
Reload-Mamba & ConvNeXt-Tiny + multi-scale decoder & SS, val & \textbf{83.2} \\
\hline
\end{tabular}
\end{table*}

Reload-Mamba reaches 83.2\% single-scale mIoU on Cityscapes, outperforming SegNeXt-B (82.6\%) and all listed ResNet-101 baselines including AlignSeg (82.4\%). Given that Cityscapes contains many thin structures (poles, traffic signs) and dense boundary regions where propagation-induced dilution is most severe, this result provides additional evidence that hierarchical anti-dilution refinement is particularly suited for boundary-sensitive urban scene segmentation. The advantage is consistent with the design intent: the boundary-supervised prior concentrates restoration on boundary regions, the class-uncertainty-aware gate further amplifies it at ambiguous pixels, and the multi-level Reload aggregates these effects across decoder scales.

\subsubsection{Results on PASCAL VOC}
We follow the standard DeepLab-style PASCAL VOC training protocol with MS-COCO pre-training, which is the common setting for ResNet-101 based VOC segmentation methods. Table~\ref{tab:comparison_voc} reports the comparison. Reload-Mamba achieves 87.8\% mIoU on the validation set with ResNet-101 + COCO pre-training, on par with EMANet~\cite{li2019emanet} (87.7\%) and DeepLabv3+ (Xception-65, 87.8\%),
and outperforming CFNet~\cite{zhang2019cfnet} (87.2\%) and DFN~\cite{yu2018dfn} (86.2\%)
under the same protocol. These results suggest that directional state-space propagation with class-uncertainty-aware Reload Gate-based local response restoration can match or surpass attention-based and dilation-based context modeling under the standard PASCAL VOC training protocol. Since most recent compact Transformer-based and Mamba-based segmentation models do not report PASCAL VOC results, we treat this comparison as a cross-protocol evaluation against established ResNet-101 based segmentation methods.

\begin{table*}[htbp]
\centering
\caption{Comparison on the PASCAL VOC 2012 benchmark. COCO denotes MS-COCO pre-training before fine-tuning on the PASCAL VOC augmented training set, which is the standard protocol for ResNet-based VOC segmentation methods. Reload-Mamba follows the same protocol on PASCAL VOC.}
\label{tab:comparison_voc}
\begin{tabular}{lcccc}
\hline
Method & Backbone & COCO & Split & mIoU (\%) \\
\hline
FCN \cite{long2015fcn} & VGG-16 & -- & test & 62.2 \\
DeepLabv2 \cite{chen2017deeplab} & ResNet-101 & -- & val & 74.9 \\
DeepLabv3 \cite{chen2017deeplab} & ResNet-101 & -- & val & 79.3 \\
DeepLabv3+ \cite{chen2018deeplabv3plus} & ResNet-50 & -- & val & 79.84 \\
DANet \cite{fu2019danet} & ResNet-101 & -- & val & 80.4 \\
\hline
DeepLabv3 \cite{chen2017deeplab} & ResNet-101 & \checkmark & val & 82.7 \\
PSPNet \cite{zhao2017pspnet} & ResNet-101 & \checkmark & test & 85.4 \\
EncNet \cite{zhang2018encnet} & ResNet-101 & \checkmark & test & 85.9 \\
DFN \cite{yu2018dfn}         & ResNet-101  & \checkmark & test & 86.2 \\
CFNet \cite{zhang2019cfnet}  & ResNet-101  & \checkmark & test & 87.2 \\
EMANet \cite{li2019emanet}   & ResNet-101  & \checkmark & test & 87.7 \\
DeepLabv3+ \cite{chen2018deeplabv3plus} & Xception-65 & \checkmark & test & 87.8 \\
RecoNet \cite{chen2020reconet} & ResNet-101 & \checkmark & test & 88.5 \\
\hline
Reload-Mamba & ResNet-101 & \checkmark & val & \textbf{87.8} \\
\hline
\end{tabular}
\end{table*}

We note that PASCAL VOC results in Table~\ref{tab:comparison_voc} include both val and test splits depending on the original protocol of each baseline, as VOC test server access is no longer available for new submissions. We therefore report Reload-Mamba on the val split and place it alongside other ResNet-101 based methods to provide a consistent cross-method comparison.

\subsection{Ablation Study}
To analyze the effectiveness of Reload-Mamba, we conduct ablation studies on the architectural progression, the cumulative contribution of the three segmentation-specific designs, the Reload Gate composition, the directional scanning strategy, the loss components, the placement of hierarchical Reload, the boundary supervision, the auxiliary head placement, the per-level channel widths, backbone compatibility, hyperparameters, and inference strategies. Unless otherwise specified, all ablation experiments are conducted on ADE20K and evaluated under single-scale inference without horizontal flipping or multi-scale testing. Therefore, the full model in the ablation study corresponds to the 47.9\% single-scale ADE20K result, while the 48.9\% result in the main comparison table is obtained with multi-scale inference and horizontal flipping.

\subsubsection{Ablation Protocol}
For a fair comparison, all ablation experiments follow the same optimization setting, data augmentation strategy, crop size, and single-scale inference protocol summarized in Table~\ref{tab:ablation_protocol}. The baseline model adopts the same encoder-decoder structure but removes the entire hierarchical Reload-Mamba module (including the boundary-supervised prior, the class-uncertainty-aware gate, and multi-level Reload) as well as the Sobel edge prior. The full model corresponds to the final Reload-Mamba architecture evaluated under single-scale inference.

\begin{table}[htbp]
\centering
\caption{Default ablation setting on ADE20K. All ablation results are reported under single-scale inference without test-time augmentation.}
\label{tab:ablation_protocol}
\begin{tabular}{lc}
\hline
Inference & Single-scale, no flip, no multi-scale \\
\hline
Dataset & ADE20K \\
Backbone & ConvNeXt-Tiny \\
Crop size & $512 \times 512$ \\
Optimizer & AdamW \\
Initial learning rate & $1 \times 10^{-4}$ \\
Epochs & 200 \\
Batch size & 4 \\
Gradient accumulation & 4 \\
\hline
\end{tabular}
\end{table}

\subsubsection{Architectural Progression of the Single-Level Reference Baseline}
We first construct an architectural progression of a single-level, implicit-prior, uncertainty-agnostic anti-dilution variant applied to semantic segmentation. This serves as the reference baseline upon which the three segmentation-specific designs of Reload-Mamba are built. Table~\ref{tab:ablation_components} shows that each conventional component (Sobel edge prior, four-directional scanning, single-level Reload Gate at $D_2$, auxiliary supervision) provides incremental gains, but the progression saturates at $45.7\%$ mIoU---the single-level reference point used in Section~\ref{sec:ablation_cumulative}.

\begin{table}[htbp]
\centering
\caption{Architectural progression of a single-level anti-dilution reference baseline on ADE20K under single-scale inference. The final row corresponds to a single-level configuration (implicit prior, no class uncertainty, refinement placement at $D_2$) and serves as the starting point of Table~\ref{tab:ablation_cumulative}.}
\label{tab:ablation_components}
\begin{tabular}{llc}
\hline
Setting & Configuration & mIoU (\%) \\
\hline
Baseline                                          & Encoder-decoder only                                  & 43.3 \\
+ Edge Prior                                      & Baseline + Sobel edge prior                            & 43.9 \\
+ Directional Scan                                & Baseline + four-directional scan                       & 44.8 \\
+ Scan + Edge Prior                               & Directional scan + Sobel edge prior                    & 45.1 \\
+ Single-level Reload Gate (no $U$, implicit prior) & Single-level refinement at $D_2$                                & 45.5 \\
+ Aux Head                                          & Single-level baseline (implicit prior, no $U$, $D_2$ placement) & \textbf{45.7} \\
\hline
\end{tabular}
\end{table}

\subsubsection{Cumulative Effect of Segmentation-Specific Designs}
\label{sec:ablation_cumulative}
The central ablation of this work is reported in Table~\ref{tab:ablation_cumulative}: we progressively add the three segmentation-specific designs on top of a single-level anti-dilution baseline ($45.7\%$) to verify that each contributes beyond what a uniform single-level refinement provides. All settings share the same backbone, decoder, four-directional scanning, and training objective except for the three components under study.

\begin{table*}[htbp]
\centering
\caption{Cumulative contribution of the three segmentation-specific designs on ADE20K under single-scale inference. The first row corresponds to a single-level anti-dilution baseline ($D_2$ placement, implicitly estimated prior, no class uncertainty in the gate).}
\label{tab:ablation_cumulative}
\begin{tabular}{lcccc}
\hline
Setting & Boundary-Sup. Prior & Class-Unc. Gate & Multi-Level Reload & mIoU (\%) \\
\hline
Single-level baseline       & --        & --        & --        & 45.7 \\
+ Boundary-Sup. Prior       & \checkmark & --        & --        & 46.3 \\
+ Class-Unc. Gate           & \checkmark & \checkmark & --        & 47.0 \\
+ Multi-Level Reload (Full) & \checkmark & \checkmark & \checkmark & \textbf{47.9} \\
\hline
\end{tabular}
\end{table*}

Table~\ref{tab:ablation_cumulative} shows that each of the three segmentation-specific designs contributes a measurable improvement beyond the single-level baseline. The boundary-supervised prior alone contributes $+0.6$ mIoU, indicating that explicit boundary localization is beneficial in multi-class segmentation even before any gate modification. Adding the class-uncertainty-aware gate provides a further $+0.7$, suggesting that the entropy signal from the pre-reload auxiliary head identifies dilution-vulnerable regions that the boundary prior alone does not capture. Finally, hierarchical multi-level Reload adds $+0.9$, confirming that propagation-induced dilution operates at multiple semantic scales and is not addressable by a single-level gate. The cumulative improvement of $+2.2$ mIoU establishes that the three designs are complementary rather than overlapping, and that a uniform single-level anti-dilution refinement leaves substantial performance on the table.

\subsubsection{Reload Gate Design}
We analyze the internal design of the Reload Gate, focusing on how the fused contextual feature $M^{(l)}$ is combined with the original decoder feature $D_l$ to produce the restored feature $D_m^{(l)}$. The uncertainty-agnostic baseline removes the class uncertainty $U^{(l)}$ from the gate input (Eq.~(\ref{eq:gate_new})), reducing the gate to the minimal formulation. We also include ablations that remove other gate components, including the prior modulation, the feature-change term, and the gate scaling $\alpha$.

\begin{table*}[htbp]
\centering
\caption{Ablation study of the Reload Gate design on ADE20K under single-scale inference. The class uncertainty column $U^{(l)}$ is the key segmentation-specific extension; removing it reduces the gate to an uncertainty-agnostic formulation. Gate w/o gate scaling means $\alpha = 1$ (no bounded restoration).}
\label{tab:ablation_reload_gate}
\begin{tabular}{lccccccc}
\hline
Setting & $D_l$ & $M^{(l)}$ & $|D_l-M^{(l)}|$ & Prior $P^{(l)}$ & $U^{(l)}$ & Scaling $\alpha$ & mIoU (\%) \\
\hline
Fused context only            & --        & \checkmark & --        & --        & --        & --        & 43.7 \\
Naive residual addition       & \checkmark & \checkmark & --        & --        & --        & --        & 45.6 \\
Uncertainty-agnostic gate     & \checkmark & \checkmark & \checkmark & \checkmark & --        & \checkmark & 47.4 \\
Gate w/o prior modulation     & \checkmark & \checkmark & \checkmark & --        & \checkmark & \checkmark & 47.0 \\
Gate w/o difference term      & \checkmark & \checkmark & --        & \checkmark & \checkmark & \checkmark & 46.8 \\
Gate w/o gate scaling $\alpha$ & \checkmark & \checkmark & \checkmark & \checkmark & \checkmark & --        & 47.6 \\
Full Reload Gate (Ours)       & \checkmark & \checkmark & \checkmark & \checkmark & \checkmark & \checkmark & \textbf{47.9} \\
\hline
\end{tabular}
\end{table*}

Table~\ref{tab:ablation_reload_gate} confirms that all gate components are necessary. Removing the class uncertainty term (``uncertainty-agnostic gate'') causes a $-0.5$ drop, the largest single-input effect among the segmentation-specific components, validating per-pixel entropy as the most impactful addition. Removing the prior modulation $P^{(l)}$ leads to a $-0.9$ drop, indicating that boundary localization remains essential even when the uncertainty signal is present. Removing the feature-change term $|D_l - M^{(l)}|$ produces the largest single-component drop ($-1.1$), confirming that the propagation-induced change provides information not captured by either the prior or the uncertainty. Removing the gate scaling $\alpha$ produces only a small drop ($-0.3$); unbounded restoration occasionally over-restores high-frequency noise, while the bounded version provides a more conservative and consistent improvement. The naive residual baseline ($45.6$), which directly adds $D_l$ and $M^{(l)}$ without any gating, demonstrates that learned gating substantially outperforms simple feature aggregation.

\subsubsection{Directional Scanning Strategy}
Since Reload-Mamba relies on directional state-space propagation, we evaluate different scanning strategies. In this experiment, the Reload Gate, local detail prior, edge prior, and training objective are kept unchanged, and only the scan directions are varied. Table~\ref{tab:ablation_direction} shows that increasing directional diversity monotonically improves segmentation quality, with the four-directional setting providing the best result.

\begin{table}[htbp]
\centering
\caption{Ablation study of directional scanning strategies on ADE20K under single-scale inference. All settings use the same Reload Gate and training objective, while only the scan directions are changed.}
\label{tab:ablation_direction}
\begin{tabular}{lcc}
\hline
Scan Strategy         & Directions               & mIoU (\%) \\
\hline
Single scan           & 1                        & 46.4 \\
Horizontal scan       & Left + Right             & 46.8 \\
Vertical scan         & Up + Down                & 46.9 \\
Two-axis scan         & Horizontal + Vertical    & 47.3 \\
Four-directional scan & Left + Right + Up + Down & \textbf{47.9} \\
\hline
\end{tabular}
\end{table}

The single-direction setting achieves 46.4\%, already a competitive number, but each additional axis provides a measurable gain ($+0.4$ for horizontal and $+0.5$ for vertical, $+0.9$ for both axes combined, and $+1.5$ for full four-directional). The asymmetry between horizontal and vertical scanning is small, consistent with the observation that natural scenes contain semantic structures along both orientations. The four-directional configuration ($47.9$) provides the strongest aggregation of structural context and is therefore adopted in the full model.

\subsubsection{Loss Function Components}
We study the effect of different loss components. In this experiment, the network architecture is fixed to the full Reload-Mamba model, and only the training objective is changed. We progressively add OHEM, Lovász-Softmax loss, boundary-weighted loss, multi-level auxiliary supervision, and prior supervision.

\begin{table*}[htbp]
\centering
\caption{Ablation study of loss function components on ADE20K. The architecture is fixed to the full Reload-Mamba model, and only the training objective is changed.}
\label{tab:ablation_loss}
\begin{tabular}{lcccccc}
\hline
Setting & CE/OHEM & Lovász & Boundary & Aux (multi-level) & Prior $\mathcal{L}_{prior}$ & mIoU (\%) \\
\hline
CE only                       & CE   & --        & --        & --        & --        & 45.2 \\
OHEM only                     & OHEM & --        & --        & --        & --        & 45.6 \\
OHEM + Lovász                 & OHEM & \checkmark & --        & --        & --        & 46.0 \\
OHEM + Lovász + Boundary      & OHEM & \checkmark & \checkmark & --        & --        & 46.6 \\
OHEM + Lovász + Bd + Aux      & OHEM & \checkmark & \checkmark & \checkmark & --        & 47.5 \\
Full objective (Ours)         & OHEM & \checkmark & \checkmark & \checkmark & \checkmark & \textbf{47.9} \\
\hline
\end{tabular}
\end{table*}

Table~\ref{tab:ablation_loss} shows that every loss component contributes positively. Replacing CE with OHEM ($+0.4$) focuses gradients on hard pixels, particularly small categories and class boundaries. Adding Lovász-Softmax ($+0.4$) explicitly aligns training with the mIoU metric. Boundary-weighted cross-entropy ($+0.6$) further amplifies the supervision near semantic transitions, complementing the boundary-supervised prior in the architecture. Multi-level auxiliary supervision ($+0.9$) is the single largest contributor, which is consistent with its role in stabilizing both the directional fusion outputs and the uncertainty estimates used inside the Reload Gate. Finally, the prior supervision loss $\mathcal{L}_{prior}$ adds another $+0.4$, confirming that explicit prior supervision is complementary to the boundary-weighted loss applied on the segmentation output.

\subsubsection{Hierarchical Multi-Level Placement}
\label{sec:ablation_multilevel}
We compare single-level and multi-level Reload mechanisms. The single-level baselines apply the Reload Gate only at a fixed decoder level, matching a single-resolution anti-dilution placement. The multi-level variants apply the Reload Gate at multiple decoder levels and aggregate the restored representations through top-down hierarchical fusion (Eq.~(\ref{eq:hier_fuse})). We additionally evaluate placing the Reload Gate after the final upsampling stage to verify that decoder-resolution refinement is preferable to post-upsampling refinement.

\begin{table}[htbp]
\centering
\caption{Ablation of single-level vs.\ hierarchical multi-level Reload on ADE20K under single-scale inference. The last row applies the Reload Gate at the full output resolution.}
\label{tab:ablation_multilevel}
\begin{tabular}{lcccc}
\hline
Setting & $D_4$ & $D_3$ & $D_2$ & mIoU (\%) \\
\hline
Single-level @ $D_4$ (Coarse)             & \checkmark & --        & --        & 46.0 \\
Single-level @ $D_3$                      & --        & \checkmark & --        & 46.5 \\
Single-level @ $D_2$ (Finest)             & --        & --        & \checkmark & 47.0 \\
Two-level ($D_3 + D_2$)                   & --        & \checkmark & \checkmark & 47.5 \\
Two-level ($D_4 + D_2$)                   & \checkmark & --        & \checkmark & 47.4 \\
Three-level (Ours)                        & \checkmark & \checkmark & \checkmark & \textbf{47.9} \\
\hline
\multicolumn{4}{l}{\emph{Post-upsampling reference}}                  & \\
After final upsampling ($H \times W$)     & --        & --        & --        & 46.4 \\
\hline
\end{tabular}
\end{table}

Table~\ref{tab:ablation_multilevel} demonstrates that hierarchical multi-level Reload is consistently superior to any single-level placement. Among single-level variants, the finest level $D_2$ achieves the highest accuracy ($47.0$), but it still falls $0.9$ mIoU behind the three-level design. Two-level variants partially recover this gap, with $D_3 + D_2$ slightly outperforming $D_4 + D_2$, consistent with the intuition that mid-scale features ($D_3$) carry more refinement-relevant information than the coarsest decoder feature ($D_4$). Operating the Reload Gate at the final upsampled resolution ($H \times W$) yields a lower accuracy ($46.4$), confirming that anti-dilution refinement is most effective at decoder feature resolutions rather than at full output resolution: at full resolution the feature map has been smoothed by upsampling, so the propagation-induced change signal is muted and the gate loses much of its discriminative power.

\subsubsection{Boundary-Supervised Local Detail Prior}
\label{sec:ablation_prior_sup}
We compare the implicit prior estimation used in our prior binarization architecture~\cite{chan2026deepmine} with the explicit boundary supervision introduced in Section~\ref{sec:prior_and_scan}. The implicit setting removes $\mathcal{L}_{prior}$ from the total loss (Eq.~(\ref{eq:total_loss})), so the prior $P^{(l)}$ is shaped only by gradients flowing back from the downstream Reload Gate and segmentation heads. The architecture is otherwise identical across settings.

\begin{table}[htbp]
\centering
\caption{Ablation of the boundary-supervised local detail prior on ADE20K under single-scale inference. Removing $\mathcal{L}_{prior}$ reproduces an implicit prior estimation in which $P^{(l)}$ is shaped only by gradients from downstream losses.}
\label{tab:ablation_prior_sup}
\begin{tabular}{lccc}
\hline
Setting & Prior Supervision & $\lambda_{prior}$ & mIoU (\%) \\
\hline
Implicit prior                          & --         & 0.0 & 47.5 \\
Explicit boundary supervision (Ours)    & \checkmark & 0.2 & \textbf{47.9} \\
\hline
\end{tabular}
\end{table}

Explicit boundary supervision yields a $+0.4$ improvement over the implicit prior. Although the implicit setting can still learn a useful prior through gradients from the segmentation losses, the explicit boundary supervision provides a stronger and more localized training signal, especially in regions where class probabilities alone are insufficient to identify boundaries (e.g., adjacent classes with similar semantic features such as ``wall'' and ``building''). The sensitivity study in Table~\ref{tab:ablation_prior_weight} indicates that the segmentation accuracy is robust within a moderate range of $\lambda_{prior} \in [0.1, 0.4]$, with performance degrading on both ends. Over-weighting ($\lambda_{prior}=0.8$) causes the prior projection to overfit to the boundary mask at the cost of feature representation quality, while $\lambda_{prior}=0$ reproduces the implicit setting.

\begin{table}[htbp]
\centering
\caption{Sensitivity of the prior supervision weight $\lambda_{prior}$ on ADE20K under single-scale inference.}
\label{tab:ablation_prior_weight}
\begin{tabular}{cc}
\hline
$\lambda_{prior}$ & mIoU (\%) \\
\hline
0.0  & 47.5 \\
0.1  & 47.7 \\
0.2  & \textbf{47.9} \\
0.4  & 47.7 \\
0.8  & 47.3 \\
\hline
\end{tabular}
\end{table}

\subsubsection{Pre-reload vs.\ Post-reload Auxiliary Head Placement}
\label{sec:ablation_aux_placement}
The class uncertainty $U^{(l)}$ used inside the Reload Gate is derived from a pre-reload auxiliary head $\psi_u^{(l)}(M^{(l)})$ attached to the directional fusion output. An alternative is to attach the auxiliary head to the post-reload feature $D_m^{(l)}$, mirroring conventional auxiliary supervision designs. However, since $D_m^{(l)}$ depends on $U^{(l)}$ through the gate, this would create a circular dependency that requires either detaching gradients on $U^{(l)}$ or iterative inference. Table~\ref{tab:ablation_aux_placement} compares the two designs.

\begin{table}[htbp]
\centering
\caption{Effect of auxiliary head placement on ADE20K. The pre-reload variant matches our final design; the post-reload variant uses gradient detachment on $U^{(l)}$ to avoid the circular dependency, which weakens the supervision signal.}
\label{tab:ablation_aux_placement}
\begin{tabular}{lcc}
\hline
Aux Head Placement & Circular Dependency & mIoU (\%) \\
\hline
Post-reload (stop-gradient on $U$) & avoided via detach   & 47.3 \\
Pre-reload (Ours)                  & avoided structurally & \textbf{47.9} \\
\hline
\end{tabular}
\end{table}

The post-reload variant requires explicit gradient detachment on $U^{(l)}$ to break the circular dependency, which weakens the supervision signal that shapes the auxiliary head and consequently degrades the quality of the uncertainty estimate. The pre-reload variant avoids this dependency by construction and produces a $+0.6$ mIoU improvement. This confirms that the placement of the auxiliary head, not just its presence, is an important design choice in class-uncertainty-aware gating.

\subsubsection{Per-Level Channel Widths in Hierarchical Reload}
\label{sec:ablation_channels}
The hierarchical Reload mechanism uses level-dependent channel widths $(c_4, c_3, c_2)$ to control the parameter overhead from Mamba blocks at $D_3$ and $D_4$. Table~\ref{tab:ablation_channels} compares several configurations.

\begin{table}[htbp]
\centering
\caption{Effect of per-level channel widths $(c_4, c_3, c_2)$ on ADE20K. Our default configuration $(192, 128, 96)$ provides the best accuracy--parameter trade-off.}
\label{tab:ablation_channels}
\begin{tabular}{lcc}
\hline
$(c_4, c_3, c_2)$ & Params (M) & mIoU (\%) \\
\hline
(96, 96, 96)    & 47 & 47.4 \\
(192, 128, 96)  & 51 & \textbf{47.9} \\
(256, 192, 96)  & 56 & 47.9 \\
(384, 192, 96)  & 62 & 48.0 \\
\hline
\end{tabular}
\end{table}

The uniform $(96, 96, 96)$ setting loses $0.5$ mIoU compared to the default, likely because $D_4$-level Mamba scans cannot capture coarse semantic context effectively with only 96 channels. Widening the upper levels beyond the default yields marginal further gains while substantially increasing parameter count. The default $(192, 128, 96)$ offers the best accuracy--parameter trade-off and is used throughout the paper.

\subsubsection{Backbone Compatibility}
To verify the compatibility of the proposed Reload-Mamba module, we evaluate it under different encoder backbones. The module only requires the original local feature and the propagated contextual feature at each decoder level, making it applicable to different encoder-decoder segmentation frameworks.

\begin{table}[htbp]
\centering
\caption{Backbone compatibility study on ADE20K under single-scale inference. ``+ Reload-Mamba'' applies our full hierarchical Reload-Mamba module (boundary-supervised prior, class-uncertainty-aware gate, multi-level Reload) on top of the encoder-decoder baseline.}
\label{tab:ablation_backbone}
\begin{tabular}{lcc}
\hline
Backbone & + Reload-Mamba & mIoU (\%) \\
\hline
ResNet-50     & --         & 42.4 \\
ResNet-50     & \checkmark & 44.7 \\
ResNet-101    & --         & 43.6 \\
ResNet-101    & \checkmark & 45.9 \\
ConvNeXt-Tiny & --         & 43.3 \\
ConvNeXt-Tiny & \checkmark & \textbf{47.9} \\
\hline
\end{tabular}
\end{table}

The Reload-Mamba module consistently improves segmentation across all three backbones, with gains of $+2.3$ on ResNet-50, $+2.3$ on ResNet-101, and $+4.6$ on ConvNeXt-Tiny. The larger gain on ConvNeXt-Tiny suggests that the module is particularly complementary to backbones with strong hierarchical multi-scale features, where the hierarchical Reload mechanism can fully exploit the rich decoder representations. This pattern is consistent with prior observations that modern hierarchical backbones benefit more from advanced refinement modules than the classical ResNet family.

\subsubsection{Hyperparameter Sensitivity}
We study the sensitivity of two key hyperparameters: the maximum Reload Gate value $\alpha$ and the low-priority response coefficient $\lambda_{low}$. The results in Table~\ref{tab:ablation_hyperparameter} show that Reload-Mamba is relatively stable around the selected settings.

\begin{table}[htbp]
\centering
\caption{Hyperparameter sensitivity study on ADE20K under single-scale inference. The full model uses $\alpha=0.7$ and $\lambda_{low}=0.3$.}
\label{tab:ablation_hyperparameter}
\begin{tabular}{lcc}
\hline
Hyperparameter Setting & Value & mIoU (\%) \\
\hline
$\alpha$ & 0.3 & 47.3 \\
$\alpha$ & 0.5 & 47.7 \\
$\alpha$ & 0.7 & \textbf{47.9} \\
$\alpha$ & 0.9 & 47.5 \\
\hline
$\lambda_{low}$ & 0.0 & 47.4 \\
$\lambda_{low}$ & 0.1 & 47.7 \\
$\lambda_{low}$ & 0.3 & \textbf{47.9} \\
$\lambda_{low}$ & 0.5 & 47.5 \\
\hline
\end{tabular}
\end{table}

\subsubsection{Inference Strategy}
Finally, we compare different inference strategies on ADE20K. All model parameters are kept unchanged, and only the test-time inference protocol is varied.

\begin{table}[htbp]
\centering
\caption{Effect of inference strategy on ADE20K. The 47.9\% single-scale result corresponds to the setting used in the ablation studies, while the 48.9\% multi-scale + flip result corresponds to the value reported in Table~\ref{tab:comparison_ade20k}.}
\label{tab:ablation_inference}
\begin{tabular}{lccc}
\hline
Inference Strategy   & Multi-scale & Flip      & mIoU (\%) \\
\hline
Single-scale         & --        & --        & 47.9 \\
Single-scale + Flip  & --        & \checkmark & 48.2 \\
Multi-scale          & \checkmark & --        & 48.5 \\
Multi-scale + Flip   & \checkmark & \checkmark & \textbf{48.9} \\
\hline
\end{tabular}
\end{table}

\subsection{Efficiency Analysis}
The full Reload-Mamba model contains approximately 51M parameters and requires approximately 215 GFLOPs for a $512 \times 512$ input under the ADE20K single-scale inference setting. Of the 51M parameters, the ConvNeXt-Tiny encoder accounts for approximately 28M, the multi-scale decoder for approximately 5M, the three-level Mamba blocks (with the channel configuration $(192, 128, 96)$) and the associated directional attention, prior projections, and Reload Gates for approximately 14M, the three pre-reload auxiliary heads and the hierarchical fusion blocks for approximately 1.5M, and the edge prior branch and segmentation head for the remaining approximately 2.5M. Compared with the direct-port baseline that uses a single-level Reload Gate at $D_2$ only ($47$M parameters, $207$ GFLOPs), the hierarchical Reload mechanism increases parameter count by approximately $9\%$ and FLOPs by approximately $4\%$ in exchange for $+2.2$ mIoU on ADE20K. Compared with ConvNeXt-T ($60$M), VMamba-T ($55$--$62$M), and MambaVision-T ($55$M), Reload-Mamba remains parameter-efficient. We do not use efficiency as the primary comparison criterion because FLOPs, latency, and memory usage can vary significantly with implementation details and hardware, but we report these numbers for completeness.

\subsection{Qualitative Results}
To further evaluate the visual behavior of Reload-Mamba, we present qualitative segmentation examples on ADE20K, Cityscapes, and PASCAL VOC 2012. Figure~\ref{fig:qualitative_ade20k} shows qualitative results on ADE20K, which serves as the main scene parsing benchmark in this work. Each ADE20K example is arranged as the input image, the ground-truth annotation, and the Reload-Mamba prediction. Since ADE20K contains diverse object categories, complex indoor layouts, and fine-grained semantic regions, these examples are used to inspect whether the proposed model can produce coherent multi-class segmentation maps while preserving local semantic structures. Figure~\ref{fig:qualitative_cityscapes} provides Cityscapes visualizations for boundary-sensitive urban scene segmentation. Each Cityscapes example shows the ground-truth annotation, the Reload-Mamba prediction, and the overlay visualization. Since Cityscapes contains thin structures, road boundaries, vehicles, poles, and pedestrians, the overlay results help inspect the spatial alignment between predicted semantic regions and boundary-sensitive urban structures. Figure~\ref{fig:qualitative_voc} provides qualitative segmentation results on PASCAL VOC 2012 for object-centric semantic segmentation, complementing the scene-centric visualizations on ADE20K and Cityscapes.

\noindent
\begin{minipage}{\linewidth}
\centering

\begin{minipage}{0.48\linewidth}
\centering
\includegraphics[width=\linewidth]{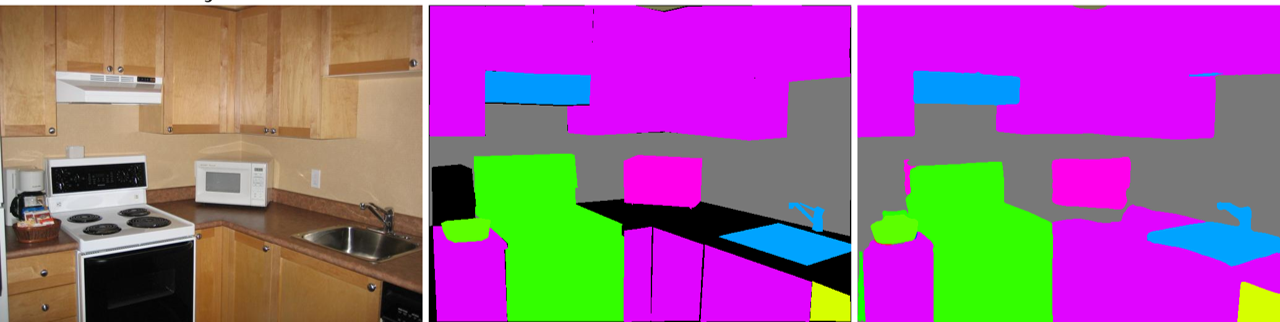}
\end{minipage}
\hfill
\begin{minipage}{0.48\linewidth}
\centering
\includegraphics[width=\linewidth]{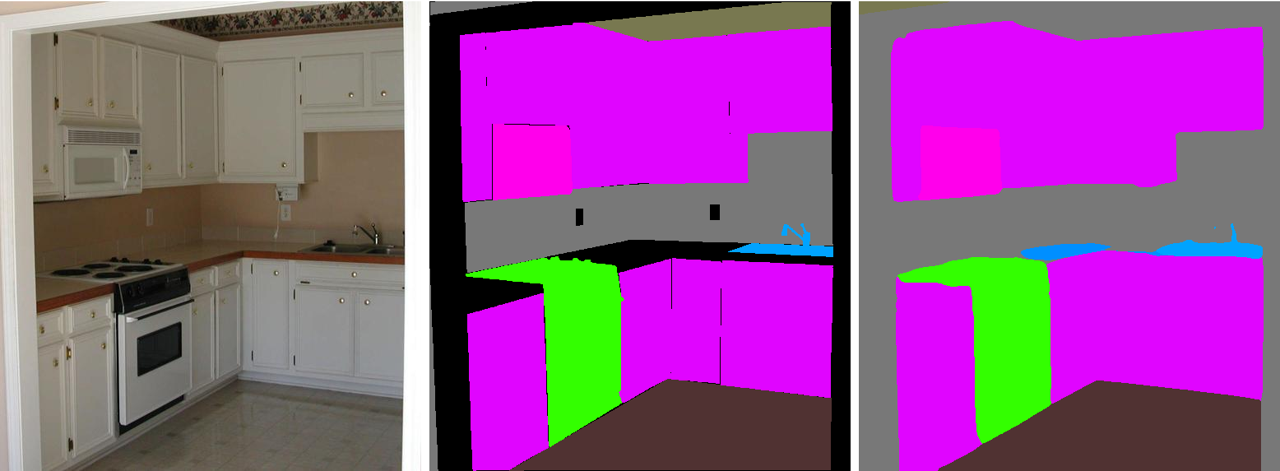}
\end{minipage}

\par\vspace{4pt}

\begin{minipage}{0.48\linewidth}
\centering
\includegraphics[width=\linewidth]{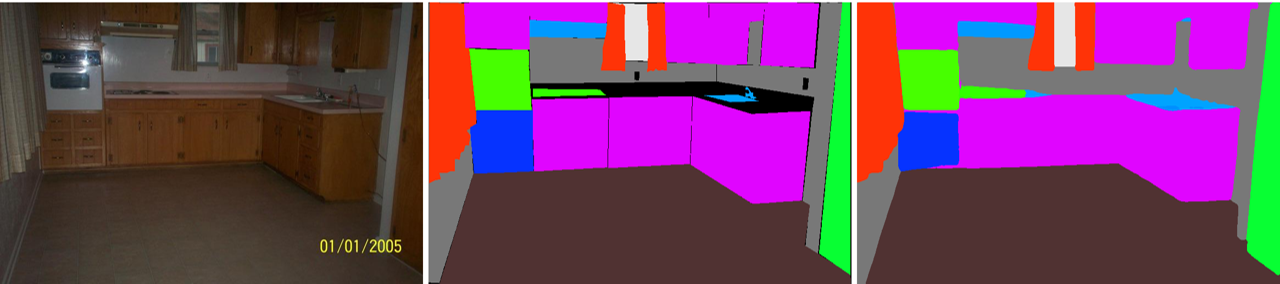}
\end{minipage}
\hfill
\begin{minipage}{0.48\linewidth}
\centering
\includegraphics[width=\linewidth]{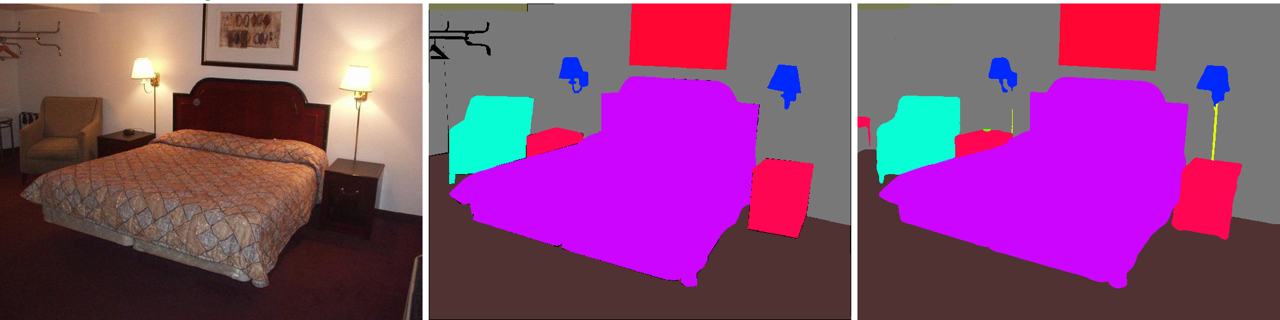}
\end{minipage}

\par\vspace{4pt}

\begin{minipage}{0.48\linewidth}
\centering
\includegraphics[width=\linewidth]{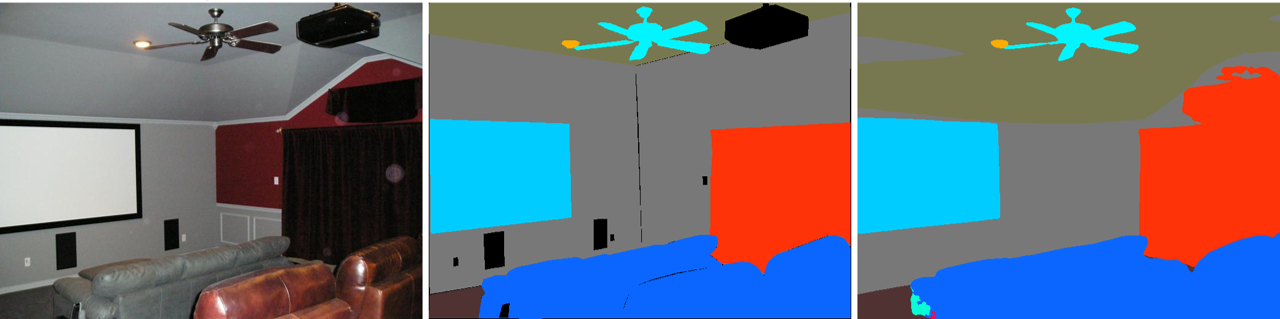}
\end{minipage}

\vspace{4pt}
\captionof{figure}{Qualitative segmentation results on ADE20K. Each example shows the input image, ground-truth annotation, and Reload-Mamba prediction.}
\label{fig:qualitative_ade20k}
\end{minipage}

\vspace{4pt}
\noindent
\begin{minipage}{\linewidth}
\centering

\begin{minipage}{0.48\linewidth}
\centering
\includegraphics[width=\linewidth]{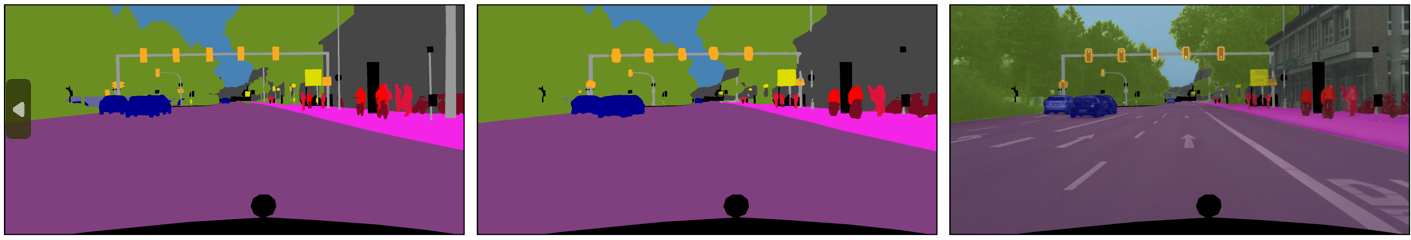}
\end{minipage}
\hfill
\begin{minipage}{0.48\linewidth}
\centering
\includegraphics[width=\linewidth]{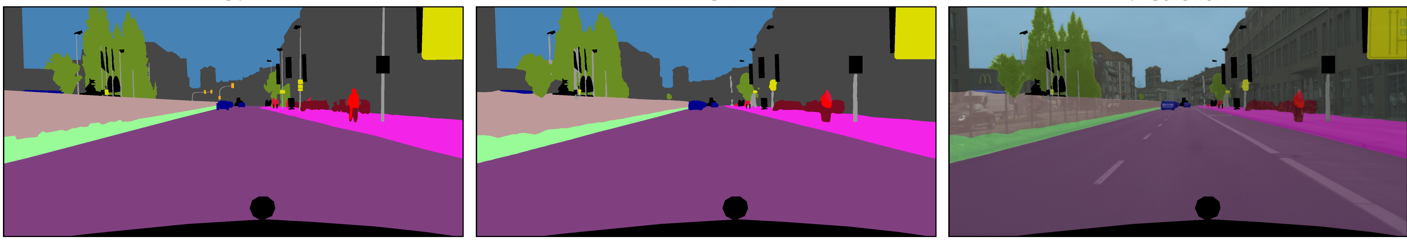}
\end{minipage}

\par\vspace{4pt}

\begin{minipage}{0.48\linewidth}
\centering
\includegraphics[width=\linewidth]{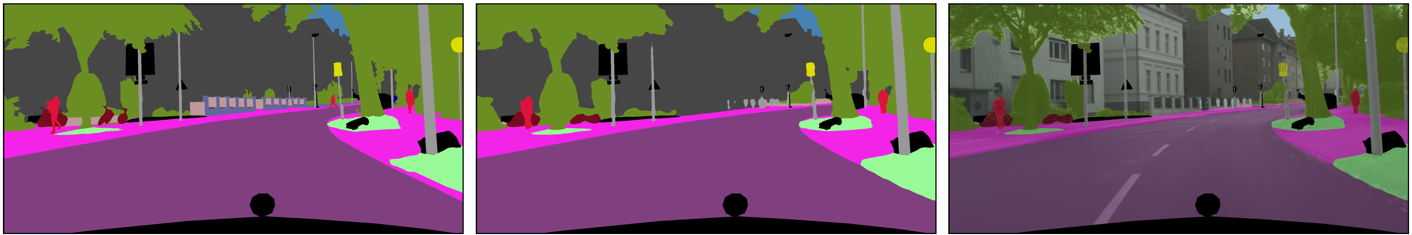}
\end{minipage}
\hfill
\begin{minipage}{0.48\linewidth}
\centering
\includegraphics[width=\linewidth]{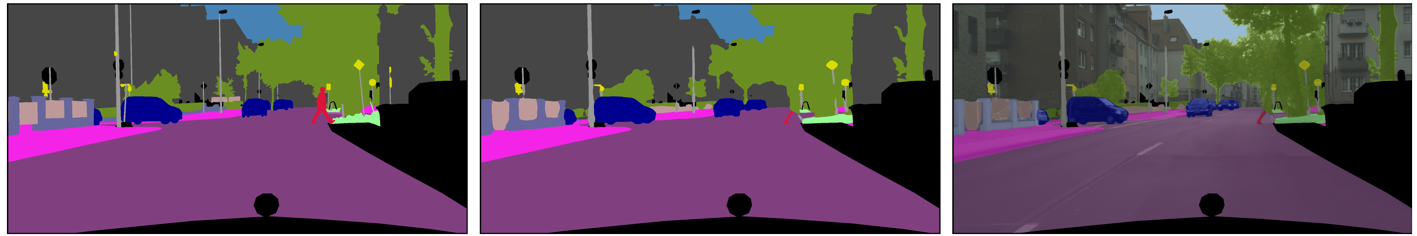}
\end{minipage}

\vspace{4pt}
\captionof{figure}{Qualitative segmentation results on Cityscapes. Each example shows the ground-truth annotation, the Reload-Mamba prediction, and the overlay visualization. The overlay results help inspect the spatial alignment between predicted semantic regions and boundary-sensitive urban structures.}
\label{fig:qualitative_cityscapes}
\end{minipage}

\vspace{4pt}
\noindent
\begin{minipage}{\linewidth}
\centering

\begin{minipage}{0.48\linewidth}
\centering
\includegraphics[width=\linewidth]{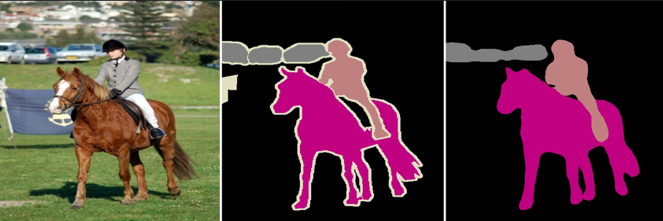}
\end{minipage}
\hfill
\begin{minipage}{0.48\linewidth}
\centering
\includegraphics[width=\linewidth]{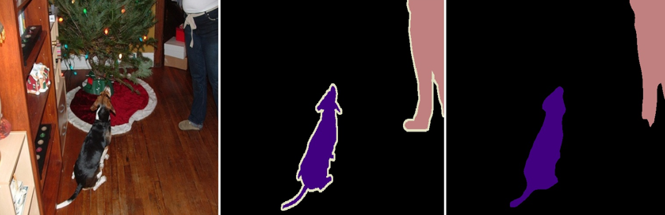}
\end{minipage}

\par\vspace{4pt}

\begin{minipage}{0.48\linewidth}
\centering
\includegraphics[width=\linewidth]{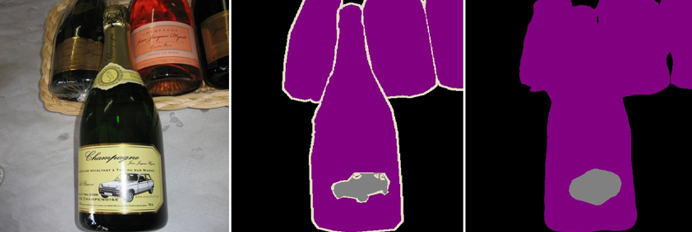}
\end{minipage}
\hfill
\begin{minipage}{0.48\linewidth}
\centering
\includegraphics[width=\linewidth]{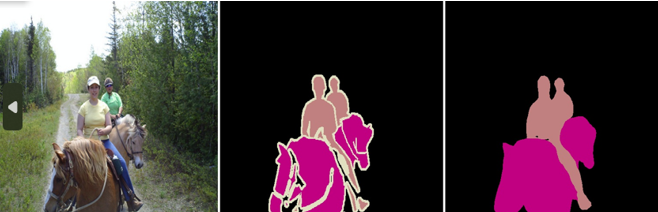}
\end{minipage}

\vspace{4pt}
\captionof{figure}{Qualitative segmentation results on PASCAL VOC 2012 for object-centric semantic segmentation.}
\label{fig:qualitative_voc}
\end{minipage}

\subsection{Limitations}
Although Reload-Mamba achieves competitive performance on multiple semantic segmentation benchmarks, this work still has several limitations. First, most ablation studies are conducted on ADE20K because it provides the most comparable setting for compact CNN-based, Transformer-based, and Mamba-based segmentation models. Although the overall trend on Cityscapes and PASCAL VOC suggests that the proposed components transfer across datasets, the per-component effectiveness on these datasets has not been systematically verified. Second, the comparison protocols across datasets are not fully symmetric: ADE20K provides relatively consistent comparisons among ConvNeXt-Tiny, SegFormer, SegNeXt, and Mamba-based models, while Cityscapes is sensitive to crop size, training split, and inference protocol, and PASCAL VOC follows the ResNet-101 + COCO-pretrain DeepLab-family setting. Therefore, the per-dataset comparisons should be interpreted within each protocol family rather than as a fully unified cross-dataset ranking. Third, the directional scan order in the current implementation is fixed to rightward, leftward, downward, and upward. Learnable or content-adaptive scan paths could potentially provide stronger directional context aggregation, but exploring such designs is beyond the scope of this work. Finally, the proposed hierarchical Reload mechanism is designed to be compatible with different Mamba-based dense prediction frameworks. However, this paper only instantiates it within a ConvNeXt-Tiny-based encoder-decoder. Its effectiveness when inserted into other visual state-space backbones such as VMamba and Vim, or applied to other dense prediction tasks such as panoptic segmentation and depth estimation, remains to be further explored.

\section{Conclusion}
In this paper, we proposed Reload-Mamba, a semantic segmentation framework that addresses propagation-induced response dilution in Mamba-based dense prediction through three segmentation-specific designs: a boundary-supervised local detail prior trained with ground-truth boundary masks, a class-uncertainty-aware Reload Gate that incorporates per-pixel class entropy from a pre-reload auxiliary head, and a hierarchical multi-level Reload mechanism that applies anti-dilution refinement at three decoder levels and fuses the restored representations top-down. Together, these designs reformulate the anti-dilution principle for multi-class dense prediction rather than reusing a direct port of the binarization architecture. Reload-Mamba achieves 47.9\% single-scale (48.9\% multi-scale) mIoU on ADE20K, 83.2\% on Cityscapes, and 87.8\% on PASCAL VOC val. Controlled ablations confirm that each of the three designs contributes beyond a direct port of the prior anti-dilution architecture, with a cumulative improvement of $+2.2$ mIoU on ADE20K. These results suggest that propagation-induced dilution in Mamba-based dense prediction is best addressed through task-specific gate conditioning, explicit prior supervision, and multi-scale refinement, rather than through direct task adaptation alone. \\

\section*{CRediT authorship contribution statement}
\textbf{Sheng-Wei Chan:} Conceptualization, Methodology, Software, Validation,
Investigation, Visualization, Writing -- original draft.
\textbf{Hsin-Jui Pan:} Software, Validation, Data curation, Investigation.
\textbf{Jen-Shiun Chiang:} Conceptualization, Supervision, Project administration,
Funding acquisition, Writing -- review \& editing.

\section*{Declaration of competing interest}
The authors declare that they have no known competing financial interests or
personal relationships that could have appeared to influence the work reported
in this paper.

\section*{Data availability}
This work uses publicly available benchmark datasets: ADE20K, Cityscapes,
PASCAL VOC 2012, and the Semantic Boundaries Dataset (SBD). No new data were
generated in this study. Source code will be made available upon reasonable
request to the corresponding author.

\section*{Declaration of generative AI and AI-assisted technologies in the writing process}
During the preparation of this work, the authors used Claude (Anthropic) to
assist with formatting checks and drafting the Highlights. After using this
tool, the authors reviewed and edited the content as needed and take full
responsibility for the content of the publication.

\noindent\textbf{Acknowledgments}\\
This research work is partially supported by National Science and Technology Council, Taiwan, under grant number: 114-2221-E-032-011-.

\bibliographystyle{cas-model2-names}
\bibliography{cas-refs}

\end{document}